\documentclass[10pt,twocolumn,letterpaper]{article}
\makeatletter
\newif\ifsupp
\@ifundefined{supp}{}{\supptrue}
\makeatother

\usepackage[]{wacv} 

\usepackage{times}
\usepackage{epsfig}
\usepackage{graphicx}
\usepackage{amsmath}
\usepackage{amssymb}
\usepackage{times}
\usepackage{microtype}
\usepackage{epsfig}
\usepackage[table,xcdraw]{xcolor}
\usepackage{caption}
\usepackage{float}
\usepackage{placeins}
\usepackage{color, colortbl}
\usepackage{stfloats}
\usepackage{enumitem}
\usepackage{tabularx}
\usepackage{xstring}
\usepackage{multirow}
\usepackage{xspace}
\usepackage{url}
\usepackage{subcaption}
\usepackage{xcolor}
\usepackage[hang,flushmargin]{footmisc}
\usepackage{graphicx}
\usepackage{amsmath}
\usepackage{amssymb}
\usepackage{booktabs}
\usepackage{arydshln}
\usepackage{amssymb}
\usepackage{pifont}
\usepackage{algorithm}
\usepackage{algpseudocode}
\usepackage{array,multirow,graphicx}

\usepackage{float}
\usepackage{amsmath}
\usepackage{enumitem}

\usepackage{threeparttable}

\usepackage[pagebackref,breaklinks,colorlinks]{hyperref}
\usepackage[accsupp]{axessibility}

\newcommand{\xmark}{\ding{55}}

\usepackage[pagebackref,breaklinks,colorlinks]{hyperref}
\usepackage[capitalize]{cleveref}
\crefname{section}{Sec.}{Secs.}
\Crefname{section}{Section}{Sections}
\Crefname{table}{Table}{Tables}
\crefname{table}{Tab.}{Tabs.}
\def\wacvPaperID{300} 
\def\confName{WACV}
\def\confYear{2024}
\begin{document}

\ifsupp
\title{Interactive Segmentation for Diverse Gesture Types Without Context - Supplementary Materials}
\else
\title{Interactive Segmentation for Diverse Gesture Types Without Context}
\fi

\author{Josh Myers-Dean{$^1$}, Yifei Fan{$^2$}, Brian Price{$^2$}, Wilson Chan{$^2$}, and Danna Gurari{$^{1,3}$}\\
{\tt\small $^1$University of Colorado Boulder $^2$Adobe Research $^3$University of Texas at Austin}}


\maketitle
\ifsupp
\section{Supplementary Materials}
 
\begin{figure*}[!b]
    \centering
    \includegraphics[width=\textwidth]{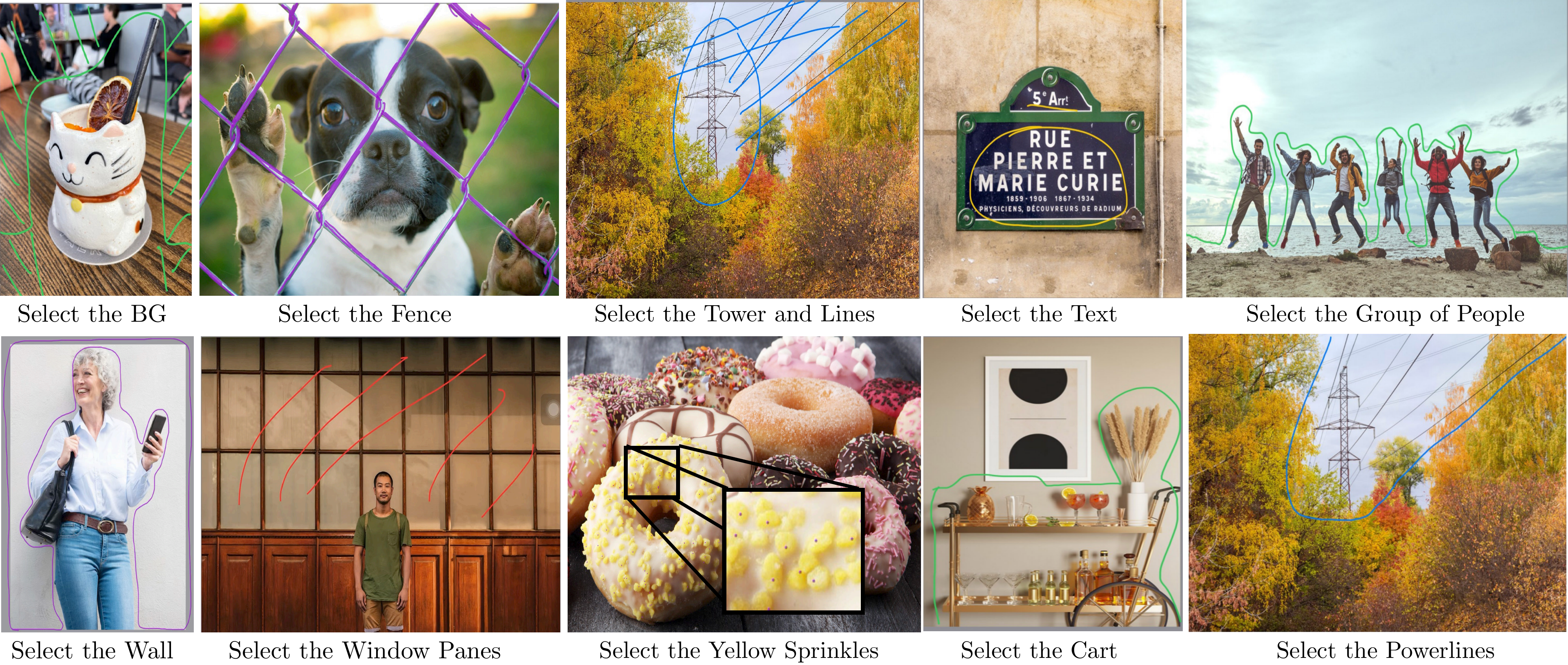}
    \caption{Examples of annotations generated in our user study about gestures. The text below each annotated image is a truncated description of the task described to the study participants. BG stands for background.}
    \label{fig: user}
\end{figure*}
This document supplements the main paper with the following:
\begin{enumerate}
    \item Alternative inputs for interactive segmentation. (supplements \textbf{Section 2})
    \item Examples of annotations from our user study. (supplements \textbf{Section 3})
    \item Examples of ground truth region segmentations that illustrate the diversity supported in our DIG dataset. (supplements \textbf{Section 4.1})
    \item Expanded discussion of the different gesture types supported in our DIG dataset. (supplements \textbf{Section 4.1})
    \item Implementation details for creating previous segmentations for our DIG dataset. (supplements \textbf{Section 4.1})
    \item Characterization of segmentations included in DIG with respect to the setting of segmentation creation versus segmentation refinement. (supplements \textbf{Section 4.2})
    \item Expanded discussion of our evaluation metric, RICE. (supplements \textbf{Sections 5 and 6})
    \item Expanded discussion of models benchmarked on our DIG dataset. (supplements \textbf{Section 6})
    \item Additional results for benchmarked models. (supplements \textbf{Section 6})
\end{enumerate}

\begin{figure}
    \centering
    \includegraphics[width=\columnwidth]{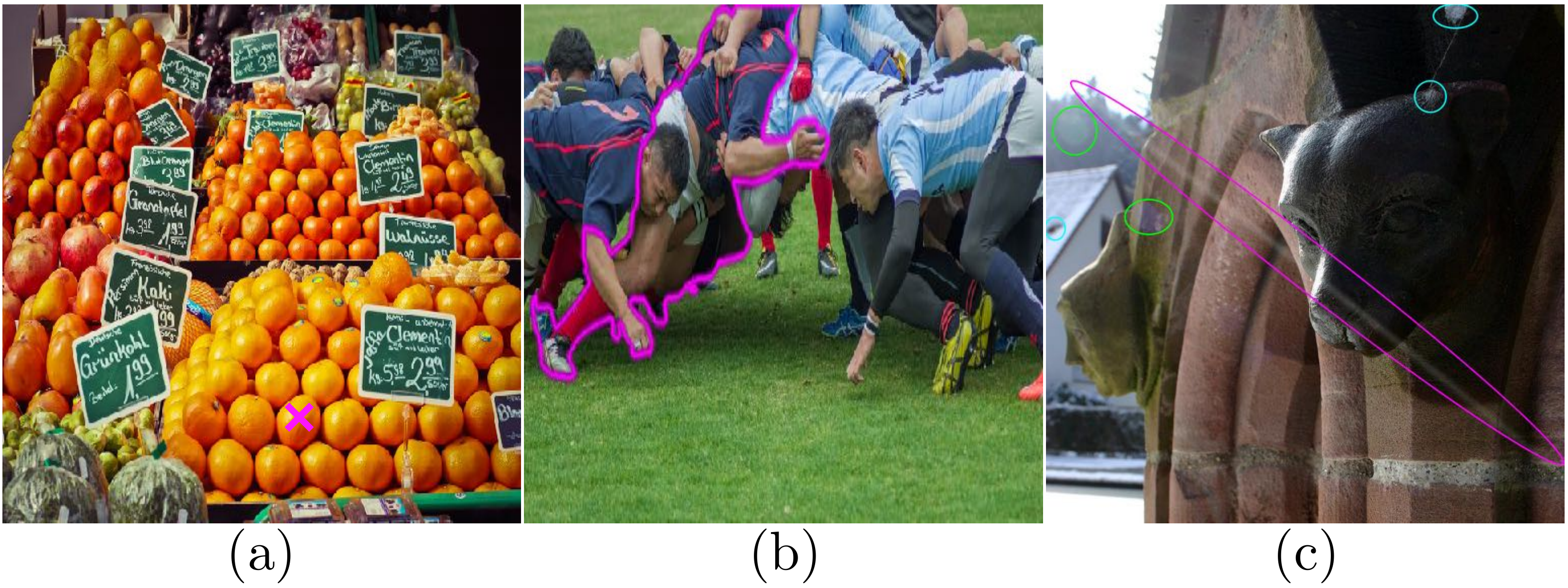}
    \caption{Scenarios representing the difficulty of using language instead of gestures: (a) selecting a specific object, (b) specifying distractors to remove from an image, (c) performing corrections on a previous segmentation.}
    \label{fig: language}
\end{figure}

\section{Non-Gesture Segmentation Methods}
Other forms of segmentation (e.g., automatic) and interaction (e.g., language) could be used to select a region in an image. With that said, we discuss here their critical shortcomings that are not present when utilizing gestures. 

For automatic methods~\cite{kirillov2019panoptic, he2017mask, long2015fully, borji2015salient}, they could allow a user to segment an entire image and then choose their selection based off of the set of segmentations. However, these methods not only lack the ability to infer user intent, as noted in the main paper, but may be unable to segment to a user's desired level of granularity (e.g., a user wants to segment a petal on a flower but a model may not support part segmentation). Furthermore, automatic methods lack the capacity for ongoing adjustments to a segmentation, as they segment an entire scene in a single operation. Conversely, the capability for users to iteratively refine segmentation until satisfaction represents a fundamental aspect of interactive segmentation. This divergence renders automatic methods inappropriate for our proposed task.

 Language models have had significant impact in many vision problems such as generation~\cite{gurari_captions, liu2023llava}, and can certainly be valuable in aiding selection~\cite{luddecke2022image, carlsson2022cross, ding2020phraseclick, kazemzadeh2014referitgame}. However, it brings important limitations.  We highlight three key limitations below.
\begin{enumerate}[itemsep=0.1em] 
\item Language models target only a tiny fraction of the 7000+ languages spoken in the world~\cite{carlsson2022cross}, and primarily focus on the most successful nations. Relying only on language would discriminate against most language groups, many of which encompass the most disenfranchised peoples, at least until vision-language models achieve the same level of accuracy for all languages as they have for English. Gestures, in contrast, can be easily understood and used by those speaking any language. 

\item Many objects/parts/regions are difficult to identify using language. To illustrate, consider three scenarios: (a) Describing which objects to manipulate is tricky as images could have repeated objects (like many oranges in a fruit stand), and users might want to adjust a subset of these objects (e.g., the magenta "X" marked orange in Figure~\ref{fig: language}(b)). The intended objects may be difficult to specify in cluttered scenes. (b) Users might struggle to articulate the desired changes to an image or region, like removing distracting color blobs (cyan-encircled areas) or unwanted elements (green-encircled lens flare) in Figure~\ref{fig: language}(b). Simply marking these would be easier than verbalizing. (c) Correcting errors from a previous segmentation could be hard to verbalize, whereas gesturing at the issue is simple. For instance, refining the player selection (e.g., removing body parts from other players) in Figure~\ref{fig: language}(c) is clearer through gestures than words.

\item Using language alone could result in an inequitable experience for disabled users. For example, typing a sentence to segment a region could result in more work for users with motor impairments. This lack of accessibility means that users could be excluded from participating in or benefiting from interactive segmentation applications that rely solely on language. In comparison, using the gesture that fits a user's unique needs (which is supported under our task) could be considerably faster and less limiting. 

\end{enumerate}

To summarize, while language is a valuable tool for computer vision tasks, it is limited in the scope of interactive segmentation. As highlighted above, language-based models are: unable to achieve comparable performance across different languages, tedious when specifying the potentially many corrections a segmentation may have, and could lead to inequitable experiences among users. In contrast, gestures are: language agnostic, trivial to specify corrections with, and can be adapted to suite the needs of users with varying speech, motor, and visual~\cite{kane2011usable} abilities.

\section{User Study Annotation Examples}
We show examples of the annotations for eight different tasks in Figure~\ref{fig: user}. These exemplify the observation in our main paper that multiple gesture types are used, with lasso the most popular (e.g., tower, powerlines, large amounts of text, group of people, car). We also found other gestures occur -- for example, we observed individuals use a combination of a lasso and scribbles to denote the background behind a cup, scribbles to select powerlines, and multiple clicks to select the many sprinkles on a donut. The diversity of gesture types underscores the need for algorithms to support multiple gesture types simultaneously.

\section{Ground Truth Segmentations of Regions}
\label{sec: gt}
To exemplify the diversity of types of regions supported in our paper, we show here examples of ground truths for (1) non-occluded regions in Figure~\ref{fig: gt_fig}(a), (2) occluded regions (i.e., \emph{parts}) in Figure~\ref{fig: gt_fig}(b), and (3) multi-region segmentations in Figure~\ref{fig: gt_fig}(c).  We also exemplify ground truth corrections for a previous segmentation in Figure~\ref{fig: train}; i.e., each correction has its own ground truth.

\begin{figure}[!h]
    \centering
    \includegraphics[width=\columnwidth]{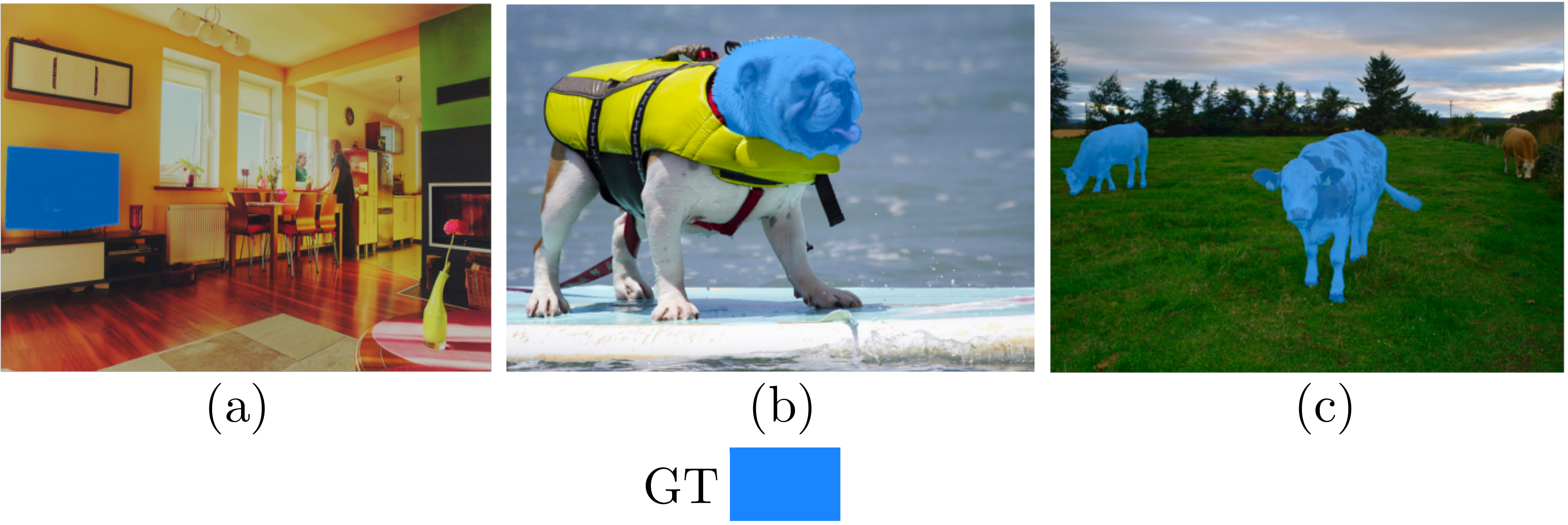}
    \caption{Examples of ground truth segmentation for multiple segmentation region types captured in our DIG dataset: (a) object (i.e., non-occluded object), (b) object part (i.e., occluded object), and (c) multi-region (i.e., multiple objects).}
    \label{fig: gt_fig}
\end{figure}

\begin{figure}[!h]
    \centering
    \includegraphics[width=\columnwidth]{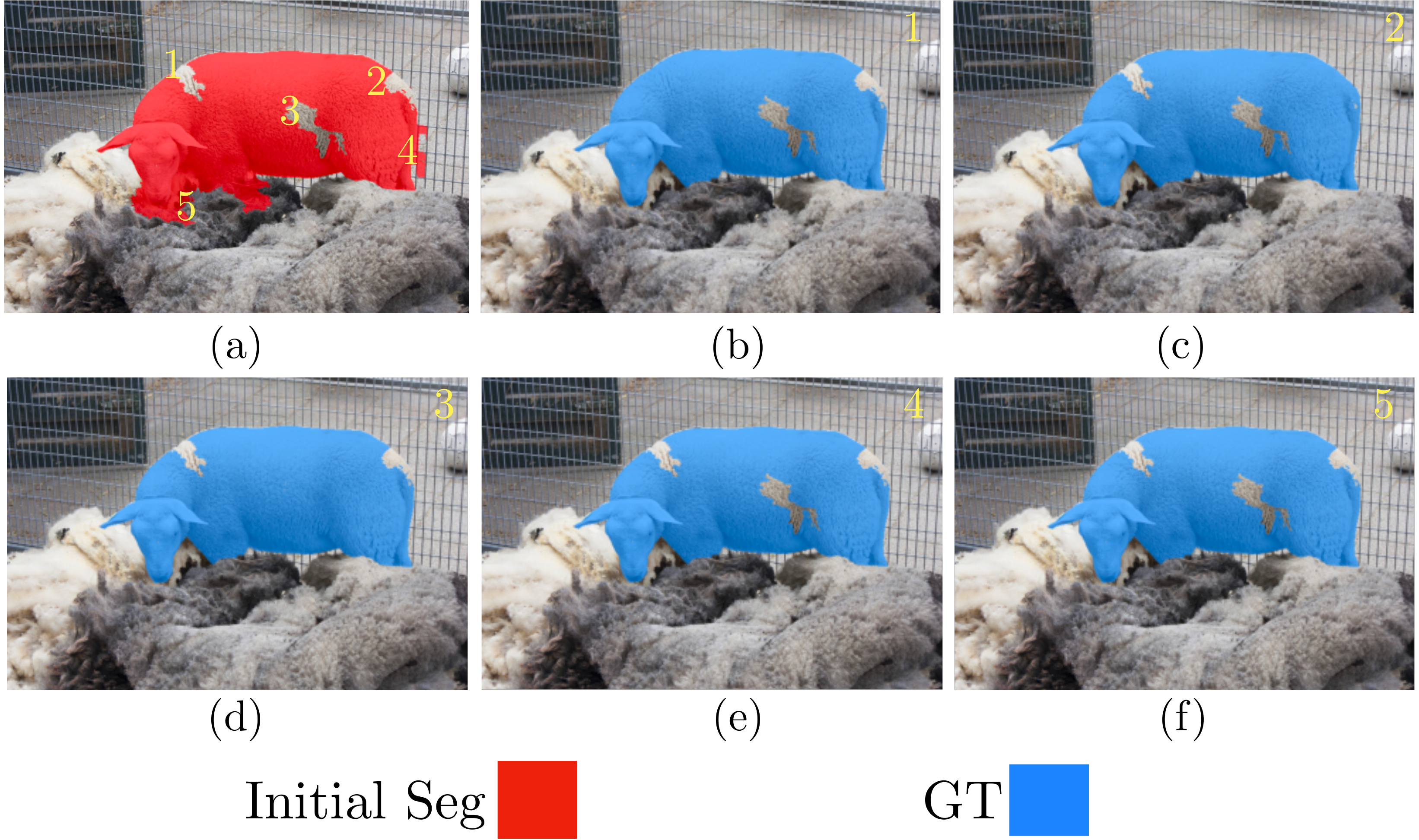}
    \caption{Examples of ground truth used for training algorithms for segmentation refinement. Each correction in an initial segmentation has a corresponding ground truth. (a) is the initial segmentation with each correction denoted with a number, (b) is the ground truth for correction \textcolor{yellow}{1}, (c) is the ground truth for correction \textcolor{yellow}{2}, and (d)-(f) are the ground truths for corrections \textcolor{yellow}{3}-\textcolor{yellow}{5}.}
    \label{fig: train}
\end{figure}

\section{Gesture Types Supported in DIG}

\subsection{Gesture Implementation Details}
For gesture annotations, we wanted a thickness that would neither be too small (i.e., a single pixel is difficult for a human to discern where the gesture is) nor too large that gestures would be indiscernible (e.g., a lasso that looks like a scribble). We chose as a heuristic a radius of 5 pixels from the center of each point in a gesture annotation. We used tools from Scikit-Image~\cite{van2014scikit} and NumPy~\cite{harris2020array} to create all markings.

\vspace{-1em}\paragraph{Lasso Generation.} We describe here how we sample points from a region boundary to create both coarse and tight lassos. To construct lassos, we uniformly sample $N$ points from the relevant boundary based on the interactive segmentation setting, randomly ``jitter" points to simulate user annotation noise, and then interpolate between the points.  For segmentation creation, the points are sampled from the ground truth segmentation of the target region.  When refining a segmentation, the points are sampled from an erroneous region (e.g., the missing leg of the turtle in Figure 1 of the main paper) such that gestures target a specific region to correct (i.e., add or subtract pixels). Coarser lassos are simulated by applying a morphological dilation operator to the boundary. Formally, let $L$ be the set of boundary points around a region and $L^\prime$ be the totally ordered set of newly sampled points. We then model the number of sampled points as $N \sim \mathcal{U}\big(\frac{|L|}{512}, \frac{|L|}{8}\big)$, where $\mathcal{U}(\cdot)$ is a discrete uniform distribution over $[a, b]$ with $a,b\in\mathbb{Z}$ and $N=|L^\prime|$. In the case of a degenerate lasso (i.e., all points in $L^\prime$ are colinear), we resample points up to 10 times. In order to simulate user noise, we randomly ``jitter" a point, $p$, in $L^\prime$ before interpolating between points. Formally, $p = p \; + \; \epsilon, \; \forall_p \in L^\prime$, where $\epsilon\sim\mathcal{U}(-J, J)$ and $J$ is our ``jitter" parameter. We use $J=4$ for loose lassos and $J=0$ for tight lassos.

\vspace{-1em}\paragraph{Scribble Generation.} We describe here how we permit scribbles to pass outside a region's boundary as well as how we simulate smoother curves. First, we create the B-spline by randomly sampling 4 to 6 $(x,y)$ pairs from a target region or previous segmentation. Then, to permit scribbles to pass outside of the target region's boundaries and to simulate simpler curves, we  perturb the sampled points by independently sorting the $x$ and $y$ coordinates before interpolating. We chose as heuristics to sort the $x$ coordinates with 30\% probability and the $y$ coordinates with 60\% probability. This allows the curves to pass outside of the boundaries of regions as the new $(x,y)$ pairs may not exist within the region. Moreover, sorted points lead to visually less complex curves.

\vspace{-1em}\paragraph{Rectangle Generation.} We describe how we augment perfect bounding boxes into diverse rectangles that encapsulate a region of interest. Given a bounding box, $B$ of the form $[x_{min}, y_{min}, x_{max}, y_{max}]$, we augment it as follows:
\begin{align}
    \label{eqn: boxes}
    x_{min/max} & = x_{min/max} + v\cdot g_i \cdot (x_{max} - x_{min})\\
    y_{min/max} & = y_{min/max} + v\cdot g_i \cdot (y_{max} - y_{min})
\end{align}
where the hyper-parameter $v$ controls the variation of the rectangle and $g_i\sim \mathcal{N}(0,1),\;\forall_i\in \{0,1,2,3\}$, where $i$ corresponds to an index (e.g., $0=y_{min}$) in $B$. We chose to randomly set $v\in [0.10, 0.15]$. Since $g_i$ can go both positive and negative, we can ``jitter" the rectangle across multiple directions.

\subsection{Gesture Timing}
We show the breakdown of how long it takes to generate each gesture type on average in Table~\ref{tab: runtimes}. We report the mean time ($\pm$ standard deviation) of each gesture type for approximately 4,740 regions (i.e., a random region from each image in the DIG validation and test splits). Computations are performed on an Intel Xeon Platinum 8275CL CPU. Overall, we observe that scribbles take approximately double the time of clicks and rectangles while lassos take 5-8 times longer. This underscores the impracticality of generating diverse gestures on the fly.

\begin{table}[!h]
\resizebox{\columnwidth}{!}{%
\begin{tabular}{clllll}
\toprule
             & Click & Scribble & Loose Lasso & Tight Lasso & Rectangle \\\midrule
Seconds & $.004\pm .001$     & $.009\pm .010$      & $.035\pm .019$ & $.023\pm .022$    & $.004 \pm .001$      \\
\bottomrule
\end{tabular}
}
\caption{Mean time $\pm$ standard deviation to generate each gesture type. We observe a large difference between the slowest gestures (i.e., lassos) and the quickest (i.e., clicks and rectangles).}
\label{tab: runtimes}
\end{table}

\subsection{Gesture Examples}
We show additional examples of gesture annotations in Figure~\ref{fig: additional_gest} to further highlight the diversity of gestures in our DIG dataset. For example, in the second row of Figure~\ref{fig: additional_gest}, we observe a nearly perfect circle lasso in (a) followed by an incomplete lasso in (b). Additionally, in the second row of Figure~\ref{fig: additional_gest}(c), we observe a scribble going out of the frame and rejoining (i.e., disconnected), contrasting the simpler scribble in the top row. Finally, we observe for clicks a diversity of positions and for rectangles varying amount of background content contained within the rectangle. The diversity of gestures present in DIG supports training models to account for the wide range of possible human interactions.
\begin{figure}[!h]
    \centering
    \includegraphics[width=\columnwidth]{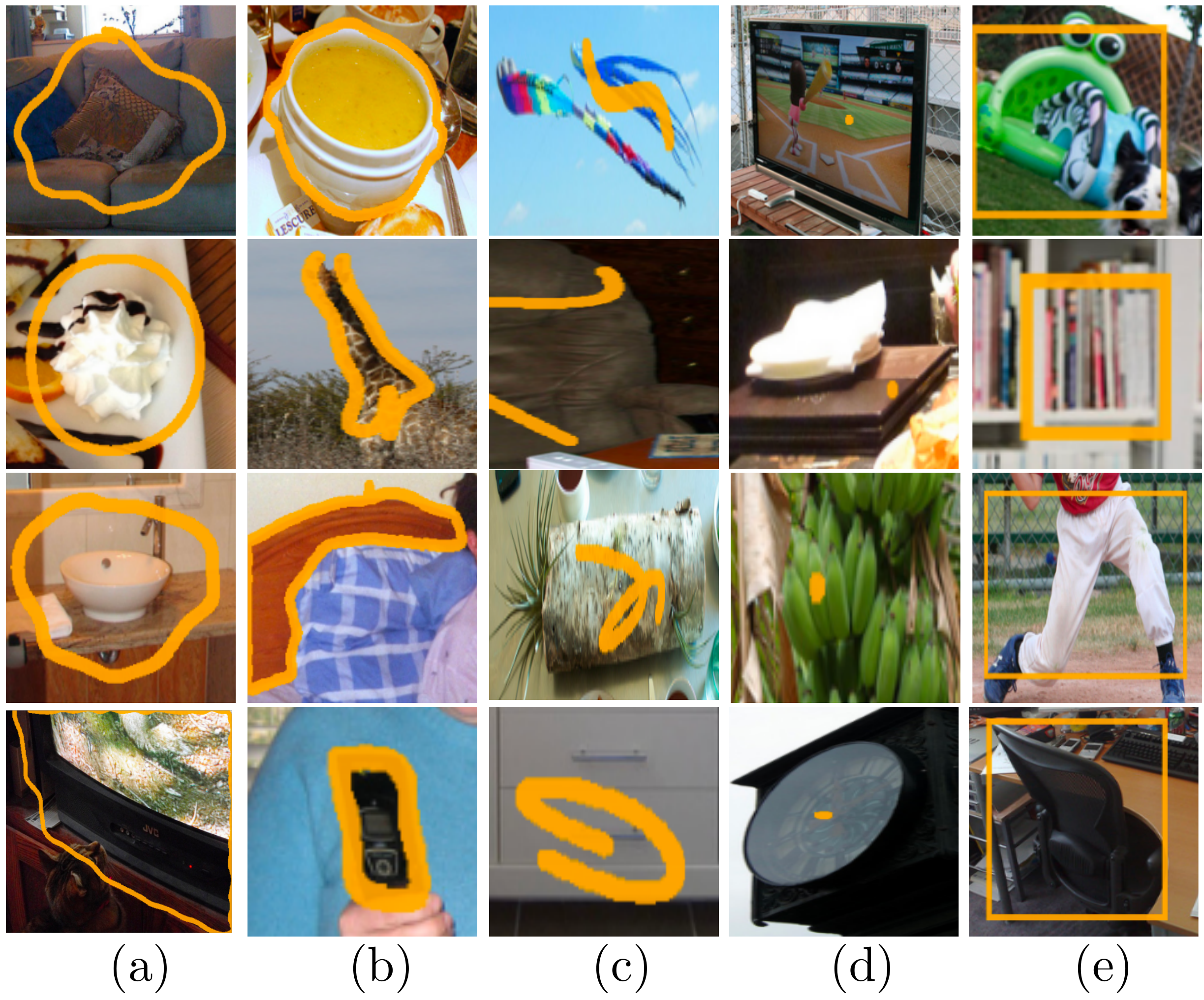}
    \caption{Examples of diversity within gesture types. We crop the image to the desired region to focus on the gesture annotations. The figure shows (a) loose lassos, (b) tight lassos, (c) scribbles, (d) clicks, and (e) rectangles.}
    \label{fig: additional_gest}
\end{figure}

\section{Previous Segmentation Construction}
For the construction of previous segmentations, we follow the approach employed by FocalClick~\cite{focalclick}. Specifically, we only retain previous segmentations results that have IoU scores between 0.75 and 0.85 with a region's ground truth.  This lower bound is motivated by prior work which shows that users of interactive segmentation methods tended to discard previous segmentations when IoU scores fell below 0.75. The upper bound is motivated by click-based segmentation methods which use 0.85 as the target IoU when evaluating using the NoC metric. 

\section{DIG Segmentation Setting Characterization}
We show the frequency of gesture types for segmentation creation and refinement in Table~\ref{tab: dig}. For the setting where no previous segmentation is present, we observe a balanced number of gesture types with a mean of 5 gestures per region. When no previous segmentation is present, we are always adding pixels to a segmentation, thus we observe no gestures intended for subtraction.  For the setting when a previous segmentation is present, there are 18.96\% fewer objects than when a previous segmentation is not present. A reason for this is that we disregard all corrections smaller than 100 pixels in areas as those corrections could be unnecessarily challenging for interactive segmentation methods. We again observe a balanced number of gesture types with a mean of 9.1 gestures per region in an image. We observe slightly more gestures whose intended context is \textit{addition} as we include multi-region segmentation scenarios, which always involve adding pixels.

 \begin{table*}[!t]
\centering
\resizebox{\textwidth}{!}{%
\begin{tabular}{cccccccccc}
\toprule
 previous segmentation  & \# Filtered Regions & \# Clicks & \# Scribbles & \# LL & \# TL & \# Rectangles & \# Subtractions & \# Additions & Mean \# Gestures per Region \\
\midrule
\xmark  & 1,091,467 & 1,091,467 & 1,091,467  & 1,082,165 & 1,085,329 & 1,091,467  & 0 & 5,411,894 & 4.98\\
$\checkmark$ & 884,551 & 1,611,618 & 1,611,618 & 1,611,506 & 1,611,578 & 1,611,618 & 4,009,524 & 4,048,382 & 9.1 \\ \midrule
\end{tabular}}
\vspace{-8pt}
\caption{Analysis of the interactive segmentation settings of segmentation creation (i.e., create segmentation from scratch) and segmentation refinement (i.e., refine a given segmentation) with respect to the frequency of different gesture types and modes. The larger number of samples for segmentation refinement stems from having multiple errors in previous segmentations needing correction. (LL = loose lasso, TL = tight lasso)}
\label{tab: dig}
\end{table*}

\section{Evaluation Metric: RICE}
\subsection{RICE Implementation}
We define RICE to take into account both the starting IoU of the previous segmentation and the ground truth, as well as the IoU of the prediction with the ground truth. Specifically, let $\hat{y}$ be the segmentation output of an interactive segmentation model, $g$ be the ground truth of a region, and $m$ be the previous segmentation. Additionally, let
\begin{equation}
IoU(A,B) = \frac{|A\cap B|}{|A\cup B|}
\end{equation}
be the intersection-over-union for some region $A$ and some ground truth $B$. Then, $\alpha = IoU(\hat{y}, g),\; \alpha \in[0,1]$ and $\beta=IoU(m, g), \; \beta\in[0,1)$ given that $\hat{y}\in\{0,1\}^{H\times W}$ is the output of an interactive segmentation method, $g\in\{0,1\}^{H\times W}$ is the ground truth for the region of interest, $m\in\{0,1\}^{H\times W}$ is an initial segmentation to refine.  We only consider initial segmentations with $\beta \in [0, 1)$ for this metric since $\beta=1$ would be a perfect segmentation with no available refinements. When creating a segmentation with no previous segmentation, RICE simplifies to $IoU(\cdot)$.

RICE ranges from a minimum value of -1 to a maximum value of 1. Intuitively, a positive score means that an algorithm corrected a previous segmentation (i.e., $\alpha > \beta$), and a score of 0 means the algorithm either did not change the previous segmentation or that the algorithm produced a similar result to the previous segmentation (i.e., $\alpha=\beta$). A negative score means that the algorithm damaged the previous segmentation (i.e., $\alpha < \beta$). This contrasts previous metrics that only range from 0 to 1 and  do not quantify if an algorithm degrades an initial segmentation.

\begin{figure}[!h]
    \centering
    \includegraphics[width=\columnwidth]{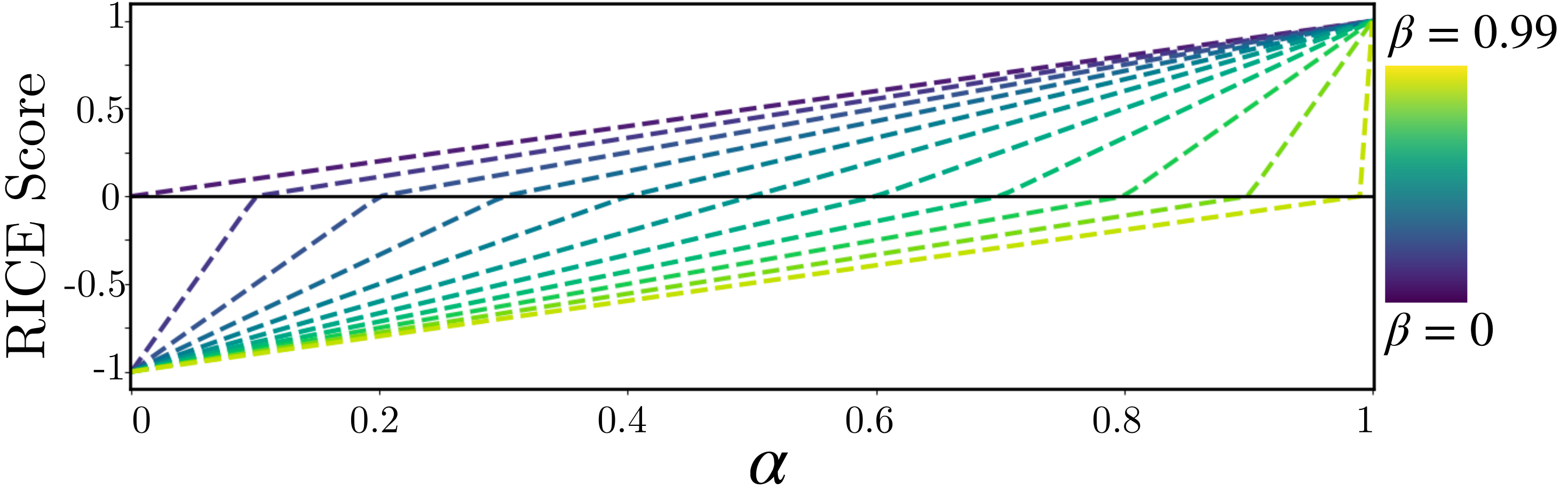}
    \caption{Visualization of the level set of RICE with fixed $\beta$ values. As $\beta$ increases so does the slope of RICE.}
    \label{fig: metric}
\end{figure}

\begin{figure*}[t!]
    \centering
    \includegraphics[width=\textwidth]{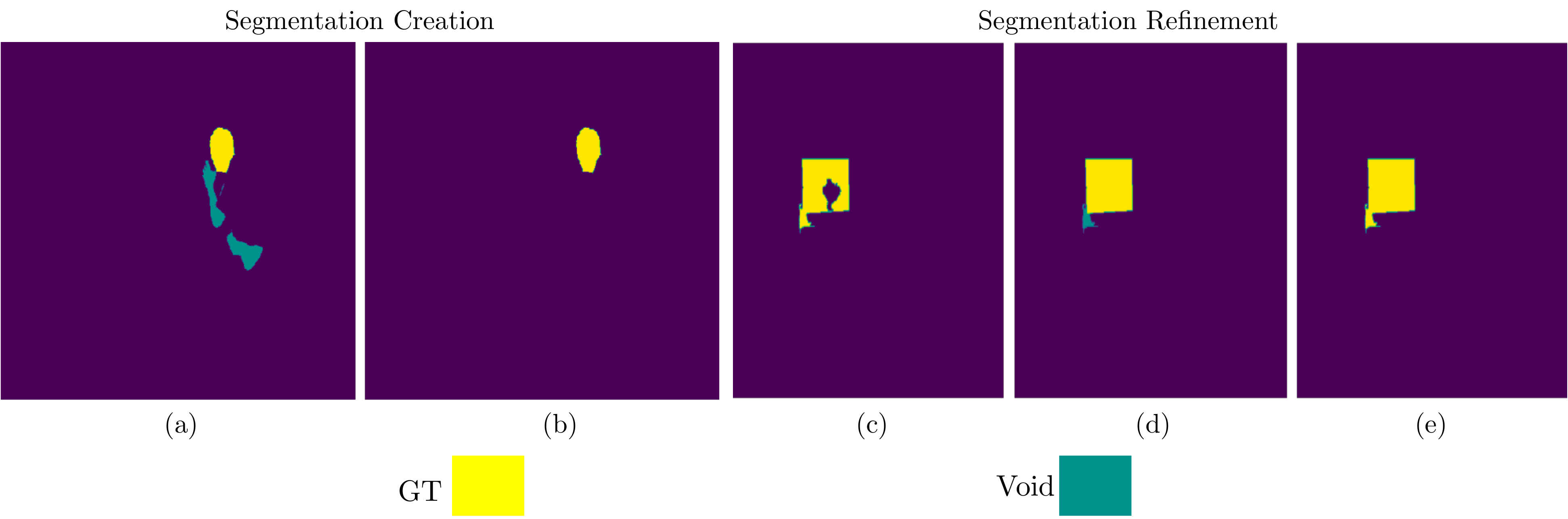}
    \caption[alt text test]{Examples of ground truths used for training (i.e., $g_a$) in a and d. These ground truths allow for the inclusion of void pixels. Examples of ground truths used for \emph{local} evaluation (i.e., $g_l$) are shown in b and e, these ground truths remove void pixels for evaluation. We show an example of a previous segmentation in (c).}
    \label{fig: gt_comb}
\end{figure*}

When examining Figure~\ref{fig: metric}, it is clear that the slope of RICE changes depending on $\beta$. Formally, we calculate the rate of change as:
\begin{align}
    \frac{\partial}{\partial \alpha}RICE(\alpha, \beta) &= \begin{cases}
  \frac{1}{1 - \beta}, & \text{if } \alpha \geq \beta\\
  \frac{1}{\beta},& \text{else}
  \end{cases}
\end{align}
First, we observe that as $\beta$ goes to 1, then the slope of RICE trends towards $\infty$ when $\alpha\geq \beta$. When $\beta$ goes to 0, the slope of RICE also trends to $\infty$ when $\alpha<\beta$. Then let, $|\alpha-\beta| = \delta$. When $\beta$ is large, then a small $\delta$ will lead to a higher RICE score than a small $\delta$ with a low $\beta$. 

To support intuitively understanding this metric, we visualize the level set of RICE in two dimensions with multiple fixed $\beta$.  Results are shown in Figure~\ref{fig: metric} with differing colors representing RICE at a fixed $\beta$. Of note, when the functions cross the $x$-axis, $\alpha=\beta$ and $RICE(\alpha, \beta)=0$. As $\beta$ increases, the slope of RICE increases, meaning a small $\delta$ will have a larger effect. To put this concept concretely, let $\beta_1=0.20$, $\beta_2=0.90$, $\alpha_1=0.21$, and $\alpha_2=0.91$ such that $|\alpha_1 -\beta_1| = |\alpha_2 - \beta_2| = \delta$. Then, RICE($\alpha_1$, $\beta_1)=0.002$ and RICE($\alpha_2$, $\beta_2)=0.10$. RICE appropriates takes into account that a small change to a relatively good previous segmentation (i.e., $\beta_2$) should signify better algorithm performance than a small change to a poor previous segmentation (i.e., $\beta_1$). Intuitively, positive changes to a relatively good previous segmentation would be more difficult (i.e., smaller areas to correct) than positive changes to a relatively poor previous segmentation.

\subsection{Local Ground Truth for Evaluation}
\label{sec: localgt}
Let $g_r\in \{0,1\}^{H\times W}$ be the entire ground truth of a given region, $g_a\in\{0,1,255\}^{H\times W}$ be the augmented ground truth from Section~\ref{subsec: dataug}, $m\in\{0,1\}^{H\times W}$ be a previous segmentation, and $g_v\in \{0,1\}^{H\times W}$ be $g_a$ with all void pixels set to 0 when creating a segmentation.  In this formulation, we define void pixels as regions in an image that do not contribute to the loss when training a model or the evaluation when evaluating a model for model selection.  Details about which pixels get labeled ``void" are provided in Section \ref{subsec: dataug}. To isolate intended parts of regions for local evaluation, we remove the void pixels for both segmentation creation and segmentation refinement. When refining a segmentation, the void pixels are set to 1 if they are an erroneous region or 0 if they are a missing region. We compute the ground truth used for local evaluation (i.e., RICE$_\text{local}$), $g_l$, as:
\begin{align}
    g_l &= (g_r \cdot m) + g_v
\end{align}
where $\cdot$ denotes element-wise multiplication. We clip the values of $g_l$ to be in $\{0,1\}$. We show comparisons of $g_l$ and $g_v$ in Figure~\ref{fig: gt_comb} for both segmentation creation and segmentation refinement.

\section{Expanded Benchmarking Discussion}

\subsection{Architecture Details for HRNet-dataAug}
For this proposed model, we exclude the post-processing module because it was designed to take in multiple clicks to locate an area intended for modification and then generate a cropped area, which could result in partially excluding the gesture and so valuable information; e.g., given that a lasso surrounds the region of interest, a crop may either not include the lasso at all, or contain parts of a region that a user does not wish to change.

\subsection{Architecture Details for Proposed Multi-Task Models}
To extend \emph{HRNet-base} to predict the intended context of a gesture in addition to the gesture type, we add in two MLPs after the encoder of HRNet~\cite{hrnet}. Specifically, after the encoder, we reduce the feature size with a $1x1$ convolution from 96 dimensions to 48 and reduce the spatial size from 128 to 32 with a max pooling operation. We then use two separate 3-layer MLPs to output a binary classification for the intended context and a 5-class classification for the gesture type. For the intended context, the MLPs go from  a dimension of 49,152 to 1024, then 1024 to 512, then 512 to 1 (i.e., add or subtract pixels) with ReLU~\cite{fukushima1975cognitron} activations between linear layers. The MLP for the gesture type follows the same architecture as the MLP for context but with an output size of 5 (i.e., one class for each gesture type). We also tested a variant that combines multi-task learning (i.e., \emph{HRNet-multiHead}) with multi-region data augmentation (i.e., \emph{HRNet-dataAug}). We follow the same training procedure as \emph{HRNet-multiHead} while employing the data augmentation described for \emph{HRNet-dataAug}.

\subsection{Data Augmentation and Training Details for Proposed Models}
\label{subsec: dataug}
 We train all variants on our DIG dataset for 15 epochs using the AdamW~\cite{loshchilov2017decoupled} optimizer with a learning rate of $3\mathrm{e}{-4}$ and batch size of 32, resize all inputs to $512\times 512$, use a value of 0.5 for non-maximal suppression, and initialize the weights using those publicly available for HRNet~\cite{hrnet} pretrained on ImageNet~\cite{imagenet}. For data augmentation, as described in Section~\ref{sec: localgt}, we define void pixels as regions that do not contribute to the loss when training a network and are ignored when evaluating models for model selection. To encourage algorithms to respond locally to user interactions, we make use of void pixels in the ground truth of our regions when applicable. For example, given that we target \emph{parts} of regions when creating a segmentation, we set the remaining connected components within a region as void. We exemplify this in Figure~\ref{fig: gt_fig}(b). Similarly, when performing refinements, we consider a specific correction targeted by an interaction as ground truth (i.e., add or subtract pixels) in addition to any part of the region that is not corrupted. We use a value of 255 to represent void pixels. We show an example of this for an initial segmentation in Figure~\ref{fig: train}.

\subsection{Input Augmentation}
\label{subsec: inputaug}
Due to existing methods only training with a small number of points (e.g., at most 24~\cite{focalclick}), they are not designed to handle the larger number of points available in some of our gesture annotations. Therefore, when applicable, we reduce the number of points in all annotations using skeletonization~\cite{zhang1984fast}.

\section{Additional Benchmarking Results}

\subsection{Deep GrabCut and IOG Results}
 We show results for both algorithms in Table~\ref{table: mulheadaug_results}. When analyzing \emph{IOG~\cite{zhang2020interactive}} and \emph{Deep GrabCut~\cite{xu2017deep}}), a plausible explanation for their worse performance is that those techniques used restrictive settings. Deep GrabCut's training process relied solely on single bounding boxes without any provision for refinement, which may have restricted the model's capability to learn general regions. Moreover, this may have limited its suitability for complex real-world settings involving small regions, multiple regions or intricate boundaries. Similarly, IOG's reliance on a unique combination of gestures, specifically bounding box and center click, may not be universally generalizable to each gesture independently, potentially limiting its practical applicability.

\subsection{Analysis for Multi-Region Segmentation}
For this task, we collect one previous segmentation region and then apply a gesture on a second disconnected region in the image to reference the second, disconnected object we also want to segment.  For ground truth, both the disconnected and the target regions of interest are needed to evaluate this set-up of multiple regions. We show results for multi-region segmentation in Table~\ref{table: mul_results}. Overall, we observe similar trends as single and multi-region segmentation with \emph{HRNet-dataAug} performing the best across the majority of gesture types and single-region methods struggling to support multiple gesture types. We observe that single-gesture methods capable of refinement (i.e., FocalClick~\cite{focalclick} and RITM~\cite{reviving2021}) perform better when using clicks, likely due to the unfair advantage we provided such methods in knowing the context of the interaction (i.e., include versus exclude annotated region). Additionally, we observe that HRNet~\cite{hrnet} with multiple classification heads and multi-region data augmentation performs the second best, further showing the efficacy of the proposed data augmentation. Finally, we observe that our other models without data augmentation (i.e., \emph{HRNet-base} and \emph{HRNet-head}) perform the worst, likely due to having neither the context of the interaction nor the multi-region setting included in training.

\begin{table*}[!t]
  \begin{threeparttable}
    \resizebox{\textwidth}{!}{\begin{tabular}{ c c c c c c c c c c c c c c c c}
    \toprule
      & \multicolumn{1}{c}{}  & \multicolumn{2}{c}{\bf Average } &\multicolumn{2}{c}{\bf Click} & \multicolumn{2}{c}{\bf Scribble} & \multicolumn{2}{c}{\bf Loose Lasso} & \multicolumn{2}{c}{\bf Tight Lasso} & \multicolumn{2}{c}{\bf Rectangle} \\
     \cmidrule(lr){3-4}
     \cmidrule(lr){5-6}
     \cmidrule(lr){7-8}
     \cmidrule(lr){9-10}
     \cmidrule(lr){11-12}
     \cmidrule(lr){13-14}
         &\bf Method &  \bf RICE$_{\text{local}}$ & \bf RICE$_{\text{global}}$  & \bf RICE$_{\text{local}}$ & \bf RICE$_{\text{global}}$ & \bf RICE$_{\text{local}}$ & \bf RICE$_{\text{global}}$ & \bf RICE$_{\text{local}}$ & \bf RICE$_{\text{global}}$ & \bf RICE$_{\text{local}}$ & \bf RICE$_{\text{global}}$ & \bf RICE$_{\text{local}}$ & \bf RICE$_{\text{global}}$ & \\
        \midrule
        \multirow{15}{*}{\rotatebox[origin=c]{90}{\bf Multi-Region}}
        & \emph{RITM\cite{reviving2021} - positive}& -5.81 & -5.78 & \bf 28.86 & \bf 28.67 &22.57 & 21.89 &-57.40 & -56.89 &0.82 & 0.78 &-23.91 & -23.35 &\\ 
        & \emph{RITM\cite{reviving2021} - negative}& -5.70 & -5.54 & -9.02 & -8.72 &-9.51 & -9.24 &-11.23 & -11.02 &-2.83 & -2.75 &4.09 & 4.05 &\\ 
        & \emph{RITM\cite{reviving2021} - random}& -5.88 & -5.78 & 9.71 & 9.75 &6.31 & 6.10 &-34.31 & -33.93 &-1.02 & -0.99 &-10.08 & -9.82 &\\\cmidrule(lr){2-16} 
        & \emph{FocalClick~\cite{focalclick} - positive}& -16.44 & -16.27 & 23.06 & 23.12 &15.95 & 15.53 &-69.75 & -69.26 &-16.44 & -16.23 &-35.05 & -34.49 &\\ 
        & \emph{FocalClick~\cite{focalclick} - negative}& -15.25 & -14.81 & -14.58 & -14.08 &-16.02 & -15.51 &-17.41 & -17.16 &-15.98 & -15.41 &-12.27 & -11.91 &\\ 
        & \emph{FocalClick~\cite{focalclick} - random}& -16.01 & -15.70 & 4.09 & 4.39 &-0.06 & -0.04 &-43.60 & -43.23 &-16.57 & -16.18 &-23.90 & -23.43 &\\ \cmidrule(lr){2-16} 
        & \emph{SAM~\cite{kirillov2023segment}-R - positive} & -22.07 & -21.84 & -12.92 & -11.16 &-30.05 & -30.40 &-38.39 & -38.15 &-14.86 & -14.97 &-14.12 & -14.52 &\\
        & \emph{SAM~\cite{kirillov2023segment}-R - negative} & -37.88 & -37.91 & -91.99 & -91.49 &-30.05 & -30.40 &-38.39 & -38.15 &-14.86 & -14.97 &-14.12 & -14.52 &\\
        & \emph{SAM~\cite{kirillov2023segment}-R - random} & -29.75 & -29.65 & -51.33 & -50.19 &-30.05 & -30.40 &-38.39 & -38.15 &-14.86 & -14.97 &-14.12 & -14.52 &\\\cmidrule(lr){2-16}  
        & \emph{SAM~\cite{kirillov2023segment}-C - positive}& -41.14 & -39.93 & -12.92 & -11.16 &-43.28 & -41.70 &-73.89 & -72.71 &-61.49 & -59.56 &-14.14 & -14.53 &\\
        & \emph{SAM~\cite{kirillov2023segment}-C - negative}& -67.30 & -66.90 & -91.99 & -91.49 &-86.74 & -86.37 &-73.21 & -72.43 &-70.44 & -69.71 &-14.14 & -14.54 &\\
        & \emph{SAM~\cite{kirillov2023segment}-C - random}& -54.16 & -53.35 & -51.81 & -50.66 &-64.96 & -63.96 &-73.79 & -72.82 &-66.10 & -64.75 &-14.14 & -14.54 &\\\cmidrule(lr){2-16} 
        & \emph{HRNet-base}& -26.25& -46.33&-65.48&-64.95&-60.05&-60.09&-47.85&-47.75&-17.25&-18.17&-40.63&-40.68&\\   
        & \emph{HRNet-dataAug}& \bf 28.57 & \bf 27.58&-2.99&-3.00&\bf24.26&\bf23.36&\bf36.91&\bf36.08&\bf53.34&\bf50.92&\bf31.31&\bf30.52&\\ 
        & \emph{HRNet-multiHead}& -54.16& -54.32&-72.32&-72.19&-60.00&-60.01&-57.23&-57.37&-29.61&-30.36&-51.66&-51.68&\\ 
        & \emph{HRNet-multiHeadAug}&23.25 & 22.33&5.53&5.24&20.73&19.95&23.6&23.08&54.16&51.55&12.22&11.81&\\ 
        
    \bottomrule\\
    \end{tabular}}
    \end{threeparttable}
    \vspace{-15pt}
    \caption{Results on the test set of DIG for multi-region segmentation.} 
    \label{table: mul_results}
\end{table*}

\subsection{IoU for Segmentation Refinement}
We show the IoU for each method capable of segmentation refinement (i.e., all methods except IOG~\cite{zhang2020interactive} and Deep GrabCut~\cite{xu2017deep}) in Table~\ref{table: iou_results}. We exclude single region results as RICE simplifies to IOU when no previous segmentation is present. Overall, we observe significantly higher scores for IoU than the corresponding RICE scores in the main paper. This demonstrates why IoU may be misleading as an evaluation metric. For example, an algorithm may receive a high IoU score despite not improving the segmentation but rather because the previous segmentation initially had a high IoU with the region ground truth. On the other hand, RICE provides a more accurate assessment by taking into account if an algorithm improved or damaged a previous segmentation.

\begin{table*}[!t]
  \begin{threeparttable}
    \resizebox{\textwidth}{!}{\begin{tabular}{ c c c c c c c c c c c c c c c c}
    \toprule
      & \multicolumn{1}{c}{}  & \multicolumn{2}{c}{\bf Average } &\multicolumn{2}{c}{\bf Click} & \multicolumn{2}{c}{\bf Scribble} & \multicolumn{2}{c}{\bf Loose Lasso} & \multicolumn{2}{c}{\bf Tight Lasso} & \multicolumn{2}{c}{\bf Rectangle} \\
     \cmidrule(lr){3-4}
     \cmidrule(lr){5-6}
     \cmidrule(lr){7-8}
     \cmidrule(lr){9-10}
     \cmidrule(lr){11-12}
     \cmidrule(lr){13-14}
         &\bf Method &  \bf IoU$_{\text{local}}$ & \bf IoU$_{\text{global}}$  & \bf IoU$_{\text{local}}$ & \bf IoU$_{\text{global}}$ & \bf IoU$_{\text{local}}$ & \bf IoU$_{\text{global}}$ & \bf IoU$_{\text{local}}$ & \bf IoU$_{\text{global}}$ & \bf IoU$_{\text{local}}$ & \bf IoU$_{\text{global}}$ & \bf IoU$_{\text{local}}$ & \bf IoU$_{\text{global}}$ & \\
        \midrule
        \multirow{15}{*}{\rotatebox[origin=c]{90}{\bf Refinement}}
        & \emph{RITM\cite{reviving2021} - positive}& 70.71 & 67.36 & 82.62 & 79.76 &81.21 & 77.96 &51.52 & 48.13 &66.04 & 62.57 &72.16 & 68.40 &\\ 
        & \emph{RITM\cite{reviving2021} - negative}& 73.32 & 70.51 & 83.74 & 80.55 &81.32 & 78.29 &66.46 & 64.12 &68.10 & 65.39 &67.00 & 64.20 &\\ 
        & \emph{RITM\cite{reviving2021} - random}& 72.00 & 68.92 & 83.16 & 80.14 &81.26 & 78.12 &59.12 & 56.24 &66.97 & 63.89 &69.50 & 66.24 &\\ \cmidrule(lr){2-16} 
        & \emph{FocalClick~\cite{focalclick} - positive}& 62.73 & 62.68 & 73.09 & 73.38 &74.18 & 74.32 &40.58 & 40.23 &60.54 & 60.44 &65.24 & 65.04 &\\ 
        & \emph{FocalClick~\cite{focalclick} - negative}& 65.50 & 65.62 & 77.90 & 78.20 &73.38 & 73.60 &55.77 & 55.68 &61.31 & 61.41 &59.17 & 59.20 &\\ 
        & \emph{FocalClick~\cite{focalclick} - random}& 64.08 & 64.10 & 75.53 & 75.83 &73.75 & 73.90 &48.03 & 47.77 &60.95 & 60.95 &62.16 & 62.05 &\\ \cmidrule(lr){2-16} 
        & \emph{SAM~\cite{kirillov2023segment}-R - positive} & 20.03 & 20.89 & 38.49 & 41.45 &6.63 & 6.69 &25.60 & 26.40 &18.87 & 19.25 &10.58 & 10.69 &\\
        & \emph{SAM~\cite{kirillov2023segment}-R - negative} & 13.29 & 13.59 & 4.78 & 4.94 &6.63 & 6.69 &25.60 & 26.40 &18.87 & 19.25 &10.58 & 10.69 &\\
        & \emph{SAM~\cite{kirillov2023segment}-R - random} & 16.94 & 17.54 & 23.01 & 24.66 &6.63 & 6.69 &25.60 & 26.40 &18.87 & 19.25 &10.58 & 10.69 &\\ \cmidrule(lr){2-16} 
        & \emph{SAM~\cite{kirillov2023segment}-C - positive} & 23.89 & 25.39 & 41.01 & 44.08 &33.07 & 35.24 &14.79 & 15.60 &20.07 & 21.41 &10.49 & 10.59 &\\
        & \emph{SAM~\cite{kirillov2023segment}-C - negative}& 8.35 & 8.56 & 4.78 & 4.94 &6.94 & 7.24 &10.11 & 10.32 &9.45 & 9.73 &10.49 & 10.59 &\\
        & \emph{SAM~\cite{kirillov2023segment}-C - random}& 16.12 & 16.98 & 22.92 & 24.58 &19.87 & 21.12 &12.57 & 13.08 &14.72 & 15.53 &10.49 & 10.59 &\\ \cmidrule(lr){2-16} 
        & \emph{HRNet-base}& 89.03&93.84&89.23&93.59&89.08&93.9&88.97&93.11&89.35&94.54&88.51&94.07&\\   
        & \emph{HRNet-dataAug}& \bf 90.41	& 94.07&\bf90.57&93.91&\bf90.38&94.18&\bf90.57&\bf93.59&\bf90.59&94.60&\bf89.92&94.06&\\ 
        & \emph{HRNet-multiHead}& 89.58 & 94.19&89.79&\bf94.05&89.74&94.25&89.61&93.55&89.76&94.85&89.01&94.27&\\ 
        & \emph{HRNet-multiHeadAug}&88.97	&\bf 94.47&89.33&94.42&89.08&\bf94.56&88.77&93.52&89.32&\bf95.24&88.33&\bf94.63&\\ 
        
    \bottomrule\\
    \end{tabular}}
    \end{threeparttable}
    \vspace{-15pt}
    \caption{IoU results on the test set of DIG for segmentation refinement.}
    \label{table: iou_results}
\end{table*}

\subsection{Results for FocalClick with Post-Processing Module}
We tested FocalClick~\cite{focalclick} with the inclusion of their proposed post-processing module when performing refinements with clicks. Overall, we observe a RICE$_\text{local}$ score of -72.52 and a RICE$_\text{global}$ score of -71.13. One plausible explanation for the comparatively lower score may lie in the module's objectives of refining probabilities. Given that our previous segmentations are binary, it is possible that the module cannot fully exploit the probabilistic information, thereby contributing to the observed outcome. For the NoG setting, we reintroduce this module as this setting allows for multiple sequential interactions.

\subsection{Results for HRNet Multi-Task Variants}
We show results for \emph{HRNet-mutliHead} and \emph{HRNet-multiHeadAug} in Table~\ref{table: mulheadaug_results}. Overall, we observe a small boost in performance when performing local refinements (i.e., RICE$_{\text{local}}$) over \emph{HRNet-dataAug} (2.51 percentage points), but worse performance across all other metrics. We also observe that \emph{HRNet-multiHead} has slightly improved performance over \emph{HRNet-base} for segmentation refinement, illustrating that the additional tasks of gesture and context classification (in \emph{HRNet-multiHead}) play a role in helping the algorithm infer the intention of a user when no context is available.  Despite these algorithms achieving top (refinement) or near-top (creation) performance, there still remains room for improvement. This indicates that our proposed dataset challenge presents a challenging, open problem for the research community.

\subsection{NoG Evaluation}
For backwards compatibility in evaluation, we examine how long an algorithm takes to reach a sufficient quality, if at all. We examine the NoG metric, as described in the main paper, for IoU thresholds of 80, 85, and 90. In line with previous research~\cite{focalclick, reviving2021, zhang2020interactive}, we also report the number of instances where an algorithm fails to achieve the specified IoU within 20 interactions. For every interaction we follow previous work~\cite{reviving2021, focalclick, xu2016deep} by always targeting the largest error. We analyze algorithms capable of refinement both settings of starting from segmentation creation and starting from an imperfect superpixel mask supplied by DAVIS585~\cite{focalclick}. For the proposed models, we analyze the performance with respect to each gesture type supplied during training, as well as the setting of starting with a tight lasso and having each subsequent interaction be a click (i.e., mixed). We omit \emph{SAM-R} and \emph{SAM-C} for this experiment and just consider SAM~\cite{kirillov2023segment} out of the box. We adopt this modification as clicks are the only supported gesture capable of indicating content to not include in the final segmentation. 

Results are shown in table~\ref{table: nog_results}. As noted in the main paper, algorithms that require context exhibit more failures than our proposed context-free models when refining a previous segmentation, while also requiring more interactions on average to achieve a specified IoU.

\paragraph{Analysis with Respect to Context Augmentation.} 
Augmenting context for existing algorithms yields flipped results for the settings of segmentation creation and refinement. For example, FocalClick~\cite{focalclick} and RITM~\cite{reviving2021} observe the best results (outside of knowing the context) when always assuming the interaction is positive during segmentation creation. Conversely, these methods yield the best results during refinement when assuming the input context is negative. A potential rationale for this observation is that these models may be adept at adding content with minimal `cruft', leading to better results with positive context. For refinement, 77.44\% of DAVIS585~\cite{focalclick} samples contain a false positive, while RITM and FocalClick fail at reaching a sufficient IoU 22.39\% to 64.79\% of the time when assuming the context is negative, indicating that resolivng the false negatives is sufficient to reach a lower IoU (i.e., 80\%) but fails as the desired quality of the segmentation increases. For SAM~\cite{kirillov2023segment}, we observe consistent results with positive context performing the best in both scenarios. This can be partially attributed to SAM exploiting click history as mentioned in the main paper. Across all methods that take in context, we observe that random context exhibits the worst
performance in the number of gestures to reach a specific IoU, while also failing the most. This observation may be attributed to the context being selected as the opposite choice of what is desired (i.e., removal when the interaction should be addition) as well as potentially undoing any progress made.

\begin{table*}[htbp]
\centering
\resizebox{\textwidth}{!}{%
\begin{tabular}{lcccccc|cccccc}
\toprule
\multicolumn{7}{c|}{\bf Creation} & \multicolumn{6}{c}{\bf Refinement} \\
\cmidrule(lr){2-7}
\cmidrule(lr){8-13}
\bf Method & \bf NoG@80 & \bf NoG@85 & \bf NoG@90 & \bf NoF@80 & \bf NoF@85 & \bf NoF@90 & \bf NoG@80 & \bf NoG@85 & \bf NoG@90 & \bf NoF@80 & \bf NoF@85 & \bf NoF@90 \\
\midrule
\emph{FocalClick~\cite{focalclick} - positive} & \bf 10.15 & \bf 11.87 & \bf 14.45 & \bf 273 & \bf 325 & 405 & 9.07 & 10.76 & 12.95 & 242 & 295 & 364 \\
\emph{FocalClick~\cite{focalclick} - negative} & - & - & - & 585 & 585 & 585 &  7.95 & 9.75 & 12.58 &  209 &  266 &  353 \\
\emph{FocalClick~\cite{focalclick} - random}& 13.31  & 14.83  & 16.67  & 353  & 404  & 469  &  \bf 7.80  & \bf 8.98  & \bf 11.48  & \bf 198  & \bf 236  & \bf 312 \\
\hdashline
\emph{RITM~\cite{reviving2021} - positive} & \bf 8.00 & \bf 9.77 & \bf 12.89 & \bf 197 & \bf 251 &  \bf352 & 6.74 & 10.26 & \bf 13.64 & 166 & 274 & 379 \\
\emph{RITM~\cite{reviving2021} - negative} & - & - & - & 585 & 585 & 585 & \bf 5.52 & \bf 9.39 &   13.83 & \bf 131 & \bf 241 & \bf 379 \\
\emph{RITM~\cite{reviving2021} - random} & 13.62  & 14.83  & 16.35  & 378  & 416  & 466  & 14.90  & 16.07  & 17.32  & 408  & 451  & 495  \\
\hdashline
\emph{SAM~\cite{kirillov2023segment} - positive} & \bf 1.69 & \bf 1.79 & \bf 1.99 & \bf 88 & \bf 114 & \bf 181 & \bf 1.78 & \bf 1.84 & \bf 2.01 & 286 & 311 & 350 \\
\emph{SAM~\cite{kirillov2023segment} - negative} & - & - & - & 585 & 585 & 585 & - & - & - & 585 & 585 & 585 \\
\emph{SAM~\cite{kirillov2023segment} - random} & 5.82  & 6.31  & 6.33  & 129  & 172  & 269  & 5.78  & 6.06  & 6.24  & \bf 157  & \bf 200  & \bf 290  \\
\hdashline
\emph{HRNet-base - clicks}&7.29 &7.08 &6.66 & 329 &434 &526 & \bf 1.09 &\bf 1.14 &\bf 1.53 & \bf 45 & \bf 73 & \bf 114  \\
\emph{HRNet-base - scribbles}  &5.23  &6.06  &7.23  & 362  &431  &511  & 4.12  &4.62  &4.92  & 205  &284  &395   \\
\emph{HRNet-base - loose lassos} & 1.96  &2.41  &2.45  & 225  &329  &453  & 3.51  &3.89  &4.08  & 362  &421  &489   \\
\emph{HRNet-base- tight lassos} &\bf 1.28  &1.30  &1.40  &\bf 85  &\bf 122  &\bf 195  & 4.30  &5.05  &6.26  & 130  &203  &331   \\
\emph{HRNet-base - rectangles} &2.85  &3.43  &4.35  & 457  &499  &553  & 2.05  &2.15  &2.46  & 407  &445  &492   \\
\emph{HRNet-base - mixed}  &1.30  &\bf 1.23  & \bf 1.30  & 108  &144  &223  &8.18  &9.70  &11.01  & 272  &409  &530   \\\hdashline
\emph{HRNet-dataAug - clicks} & 6.99 & 7.26 & 7.40 & 202 & 309 & 456 & \bf 1.06 & \bf 1.17 & \bf 1.42 & \bf 36 & \bf 50 & \bf 88 \\
\emph{HRNet-dataAug - scribbles}  & 5.52  & 6.20  & 6.69  & 239  & 333  & 454  & 2.58  & 3.20  & 3.89  & 94  & 164  & 266  \\
\emph{HRNet-dataAug - loose lassos}  & 2.04  & 2.38  & 2.42  & 182  & 273  & 399  & 2.06& 2.44  & 2.78  & 119  & 186  & 298  \\
\emph{HRNet-dataAug - tight lassos}  & \bf 1.30  & \bf 1.32  & \bf 1.34  & \bf 69  & \bf 109  & \bf 177  & 3.36 & 4.22  & 5.33  & 99  & 178  & 283  \\
\emph{HRNet-dataAug - rectangles}  & 3.80  & 3.94  & 4.12  & 350  & 420  & 493  & 1.72 & 2.09  & 2.44  & 107  & 175  & 271  \\
\emph{HRNet-dataAug - mixed}  & 1.40  & 1.44  & 1.29  & 88  & 122  & 195  & 2.76  & 3.87  & 5.23  & 72  & 150  & 340  \\\hdashline
\emph{HRNet-multiHead - clicks} &9.82 &10.93 &13.16 & 248 &347 &491 & \bf 1.05 &\bf 1.12 &\bf 1.47 & \bf 38 &\bf 55 &\bf 89 \\
\emph{HRNet-multiHead - scribbles}  &7.82  &8.53  &10.03  & 261&356&463& 2.97&3.47&3.90& 141 &222 &333  \\
\emph{HRNet-multiHead- loose lassos}  &3.49  &3.89  &3.99  & 251&347&459& 3.67&4.07&4.05& 237 &311 &427 \\
\emph{HRNet-multiHead - tight lassos} &\bf 1.34  & \bf 1.38  &\bf 1.62  & \bf 78&\bf 110&\bf 187& 2.87&3.08&3.70& 145 &243 &345  \\
\emph{HRNet-multiHead - rectangles}  &5.44  &6.50  &8.09  & 430&490&546& 2.60&2.71&2.83& 242 &318 &398  \\
\emph{HRNet-multiHead - mixed}&1.51  &1.69  &1.66  & 89&128&208 &10.15&10.84&11.34& 224 &325 &464 \\
\hdashline
\emph{HRNet-multiHeadAug - clicks} & 10.17 & 11.47 & 13.82 & 236 & 332 & 481& \bf 1.05 & \bf 1.12 & \bf 1.47 & \bf 38 & \bf 55 & \bf 89 \\
\emph{HRNet-multiHeadAug - scribbles}  & 7.23  & 8.76 & 10.64 & 266  & 351 & 459 & 3.19 & 3.77 & 3.96 & 130 & 213 & 328 \\
\emph{HRNet-multiHeadAug - loose lassos}  & 3.64  & 3.83 & 4.05 & 255 & 350 & 455 & 3.62 & 4.13 & 4.30 & 239 & 318 & 425 \\
\emph{HRNet-multiHeadAug - tight lassos}  & \bf 1.36 &\bf 1.37 & \bf 1.70 & \bf 78 & \bf 110 & \bf 185 & 2.92 & 3.24 & 4.15 & 139 & 248 & 338 \\
\emph{HRNet-multiHeadAug - rectangles}  & 5.64 & 6.85 & 8.11 & 442 & 486 & 538 & 2.82 & 2.99 & 3.16 & 249 & 305 & 397 \\
\emph{HRNet-multiHeadAug - mixed}  & 1.55 & 1.71 & 1.66 & 93 & 130 & 204 & 10.39 & 11.10 & 12.44 & 216 & 315 & 451 \\

\bottomrule
\end{tabular}}
\caption{Algorithmic benchmarking results on DAVIS585~\cite{focalclick}. We report the number of gestures to reach a given IoU as well as the number of failures. Best results for each method are in bold.}
\label{table: nog_results}
\end{table*}

\begin{table*}[!t]
  \begin{threeparttable}
    \resizebox{\textwidth}{!}{\begin{tabular}{ c c c c c c c c c c c c c c c c}
    \toprule
      & \multicolumn{1}{c}{}  & \multicolumn{2}{c}{\bf Average } &\multicolumn{2}{c}{\bf Click} & \multicolumn{2}{c}{\bf Scribble} & \multicolumn{2}{c}{\bf Loose Lasso} & \multicolumn{2}{c}{\bf Tight Lasso} & \multicolumn{2}{c}{\bf Rectangle} \\
     \cmidrule(lr){3-4}
     \cmidrule(lr){5-6}
     \cmidrule(lr){7-8}
     \cmidrule(lr){9-10}
     \cmidrule(lr){11-12}
     \cmidrule(lr){13-14}
         &\bf Method &  \bf RICE$_{\text{local}}$ & \bf RICE$_{\text{global}}$  & \bf RICE$_{\text{local}}$ & \bf RICE$_{\text{global}}$ & \bf RICE$_{\text{local}}$ & \bf RICE$_{\text{global}}$ & \bf RICE$_{\text{local}}$ & \bf RICE$_{\text{global}}$ & \bf RICE$_{\text{local}}$ & \bf RICE$_{\text{global}}$ & \bf RICE$_{\text{local}}$ & \bf RICE$_{\text{global}}$ & \\
        \midrule
        \multirow{15}{*}{\rotatebox[origin=c]{90}{}}
        & \emph{GrabCut~\cite{xu2017deep}} &10.23&10.60& 16.21 & 16.51 & 11.66 & 12.09 & 6.16 & 6.48 & 9.01 & 9.49 & 8.13 & 8.44 &\\ 
         & \emph{IOG~\cite{zhang2020interactive}} & 13.68&13.64&23.72&23.12&19.12&19.05&3.59&3.73&12.68&12.89&9.29&9.40 &\\
         & \emph{HRNet-multiHead} &\bf 63.84 &\bf 61.28& \bf 55.20& \bf52.32& \bf57.22& \bf 54.70&\bf 66.48&\bf 64.89&\bf 81.93&\bf 78.16&\bf 58.37&\bf 56.34&\\
        & \emph{HRNet-multiHeadAug}& 58.85& 56.46&48.91&46.26&53.87&51.48&60.63&59.19&78.96&75.29&51.86&50.07&\\ \hdashline
        \multirow{6}{*}{\rotatebox[origin=c]{90}{}}
        & \emph{HRNet-multiHead}&38.55&\bf 45.86&36.39&\bf 46.68&37.84&\bf 46.46& 38.20& \bf 46.38&42.43&\bf 46.53&37.90&\bf43.24&\\
        & \emph{HRNet-multiHeadAug}&\bf 41.06& 43.61& \bf38.90&44.49& \bf 40.35&43.36&\bf 39.33&43.20&\bf45.54&44.41&\bf 41.20&40.38&\\ 
        
    \bottomrule\\
    \end{tabular}}
    \end{threeparttable}
    \vspace{-15pt}
    \caption{Results on the test set of DIG. Above the dashed line represents segmentation creation, and below represents segmentation refinement.} 
    \label{table: mulheadaug_results}
\end{table*}
\subsection{Qualitative Results}
We show qualitative results for segmentation creation when the entire region is the target in Figure~\ref{fig: nomask}, segmentation creation when the region \emph{part} is the target in Figure~\ref{fig: parts}, segmentation refinement in Figure~\ref{fig: refs}, and multi-region segmentation in Figure~\ref{fig: muls}.

In the context of creating segmentations where the target region encompasses the entire area (Figure~\ref{fig: nomask}), all of the proposed multiple-gesture variants demonstrate an inclination towards selecting an object region when provided with minimal guidance in the form of clicks and scribbles. This tendency may be attributed to the presence of void pixels in the training data, which aids in the development of algorithms that respond to local interactions as opposed to learning to identify the entirety of the region. Conversely, when the same degree of guidance is provided, techniques that employ a single gesture exhibit a proclivity towards selecting the entirety of the region. For the SAM~\cite{kirillov2023segment} variants, we observe a bias of selecting the entire player for all gesture types aside from scribbles. A potential reason for this is that for consistent evaluation, we always pick the output mask with the highest IoU with the ground truth. 

Regarding the task of segmentation creation where the target region is a \emph{part} of a region, the majority of methods tend to appropriately select the upper half of the surfer, except for Deep GrabCut~\cite{xu2017deep}, which includes the majority of the surfer as well as the board in Figure~\ref{fig: parts}. As multiple-gesture methods receive more refined guidance (e.g., lassos, rectangles), they tend to select a larger portion of the target region, thus improving the segmentation accuracy. On the other hand, the performance of single-gesture methods is suboptimal when utilizing gestures other than clicks.

In the context of segmentation refinement, our findings suggest that the multiple-gesture variants exhibit better performance in filling the missing region of the elephant while leaving other corrections relatively untouched. Conversely, single-gesture methods are observed to be ineffective in addressing the region of interest, as evidenced by FocalClick~\cite{focalclick} filling in missing content on the ear of the elephant in the second row of Figure~\ref{fig: refs}(b), instead of the intended target region on the body. Moreover, all SAM~\cite{kirillov2023segment} variants fill in the majority of corrections. We also note that single-gesture methods tend to degrade the segmentation rather than improve it when utilizing gestures other than clicks, as illustrated by the results in the first and second rows of Figure~\ref{fig: refs}(d).

In the realm of multi-region segmentation, our observations suggest that single-gesture methods methods that are provided with contextual information or trained using multi-region augmentation techniques (i.e., \emph{HRNet-dataAug, HRNet-multiHeadAug}) exhibit a superior ability to maintain the previous segmentation, which is a disjoint region, while accurately segmenting a new region of interest. We observe different fail cases for the remaining methods (i.e., \emph{HRNet-base}, \emph{HRNet-multiHead}, SAM~\cite{kirillov2023segment} variants). For example, the SAM~\cite{kirillov2023segment} variants segment only the cat while ignoring the previously segmented region.
\begin{figure*}[!b]
    \centering
    \includegraphics[width=2\columnwidth]{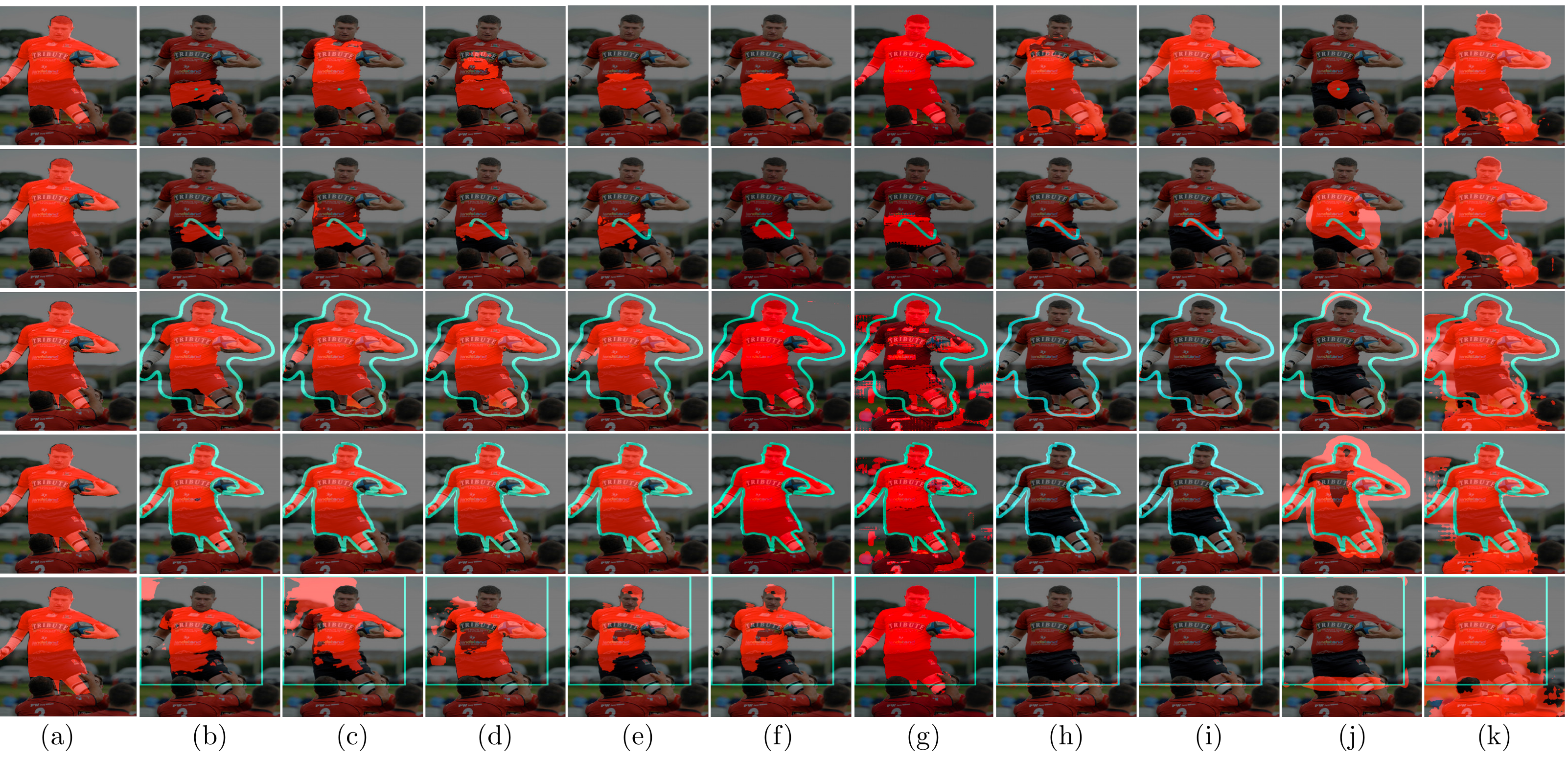}
    \caption{Qualitative results for each gesture type for segmentation creation for all methods when the target is the entire region. From top to bottom: click, scribble, loose lasso, tight lasso, rectangle. (a) input image with region ground truth overlayed, (b) \emph{HRNet-multiHeadAug}, (c) \emph{HRNet-multiHead}, (d) \emph{HRNet-dataAug}, (e) \emph{HRNet-base}, (f) \emph{SAM~\cite{kirillov2023segment}-R - positive}, (g) \emph{SAM~\cite{kirillov2023segment}-C - positive}, (h) \emph{FocalClick~\cite{focalclick} - positive}, (i) \emph{RITM~\cite{reviving2021} - positive}, (j) \emph{IOG~\cite{zhang2020interactive}}, (k) \emph{Deep GrabCut~\cite{xu2017deep}}.}
    \label{fig: nomask}
\end{figure*}

\begin{figure*}[!bth]
    \centering
    \includegraphics[width=2\columnwidth]{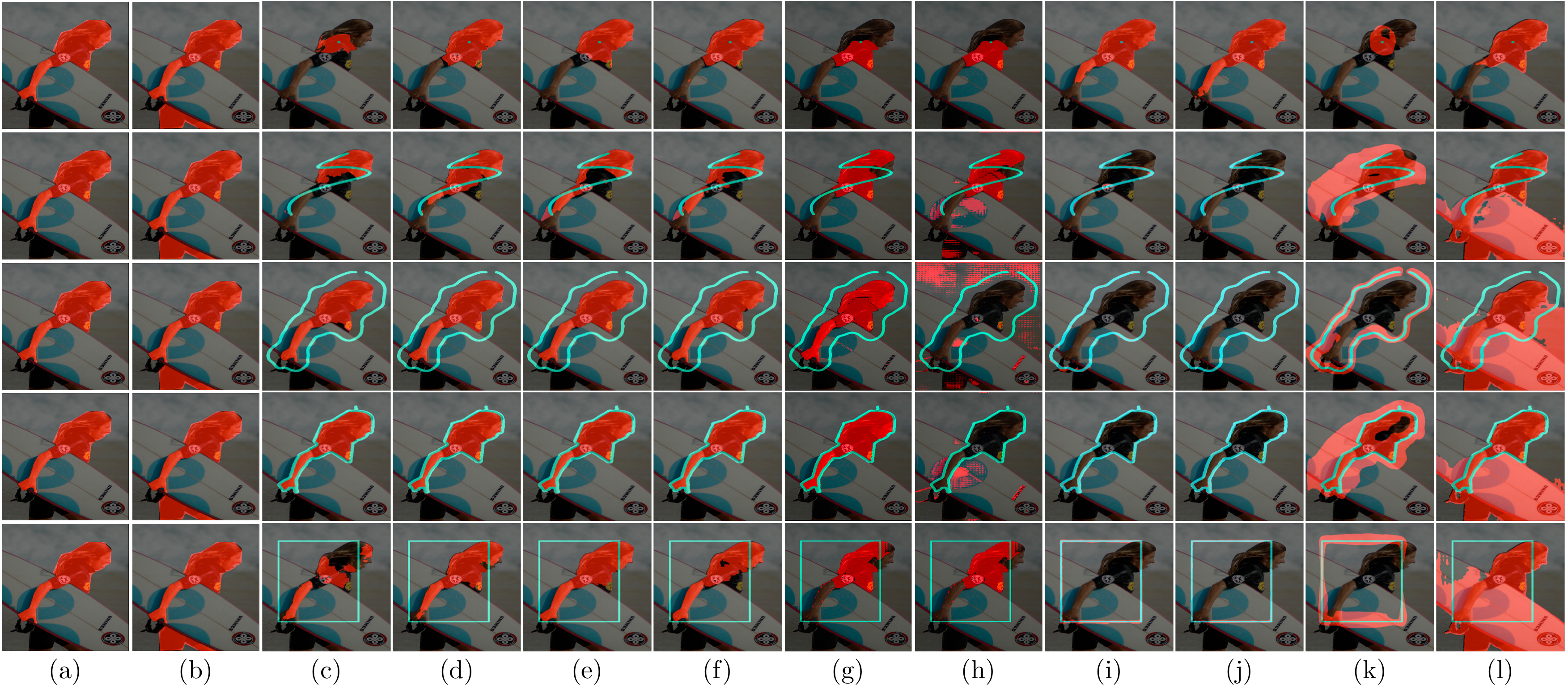}
    \caption{Qualitative results for each gesture type for segmentation creation for all methods when the target is a region \emph{part}. From top to bottom: click, scribble, loose lasso, tight lasso, rectangle. (a) input image with region ground truth overlayed, (b) input with with region part ground truth overlayed (c) \emph{HRNet-multiHeadAug}, (d) \emph{HRNet-multiHead}, (e) \emph{HRNet-dataAug}, (f) \emph{HRNet-base}, (g) \emph{SAM~\cite{kirillov2023segment}-R - positive}, (h) \emph{SAM~\cite{kirillov2023segment}-C - positive},  (i) \emph{FocalClick~\cite{focalclick} - positive}, (j) \emph{RITM~\cite{reviving2021} - positive}, (k) \emph{IOG~\cite{zhang2020interactive}}, (l) \emph{Deep GrabCut~\cite{xu2017deep}}.}
    \label{fig: parts}
\end{figure*}

\begin{figure*}[!bth]
    \centering
    \includegraphics[width=2\columnwidth]{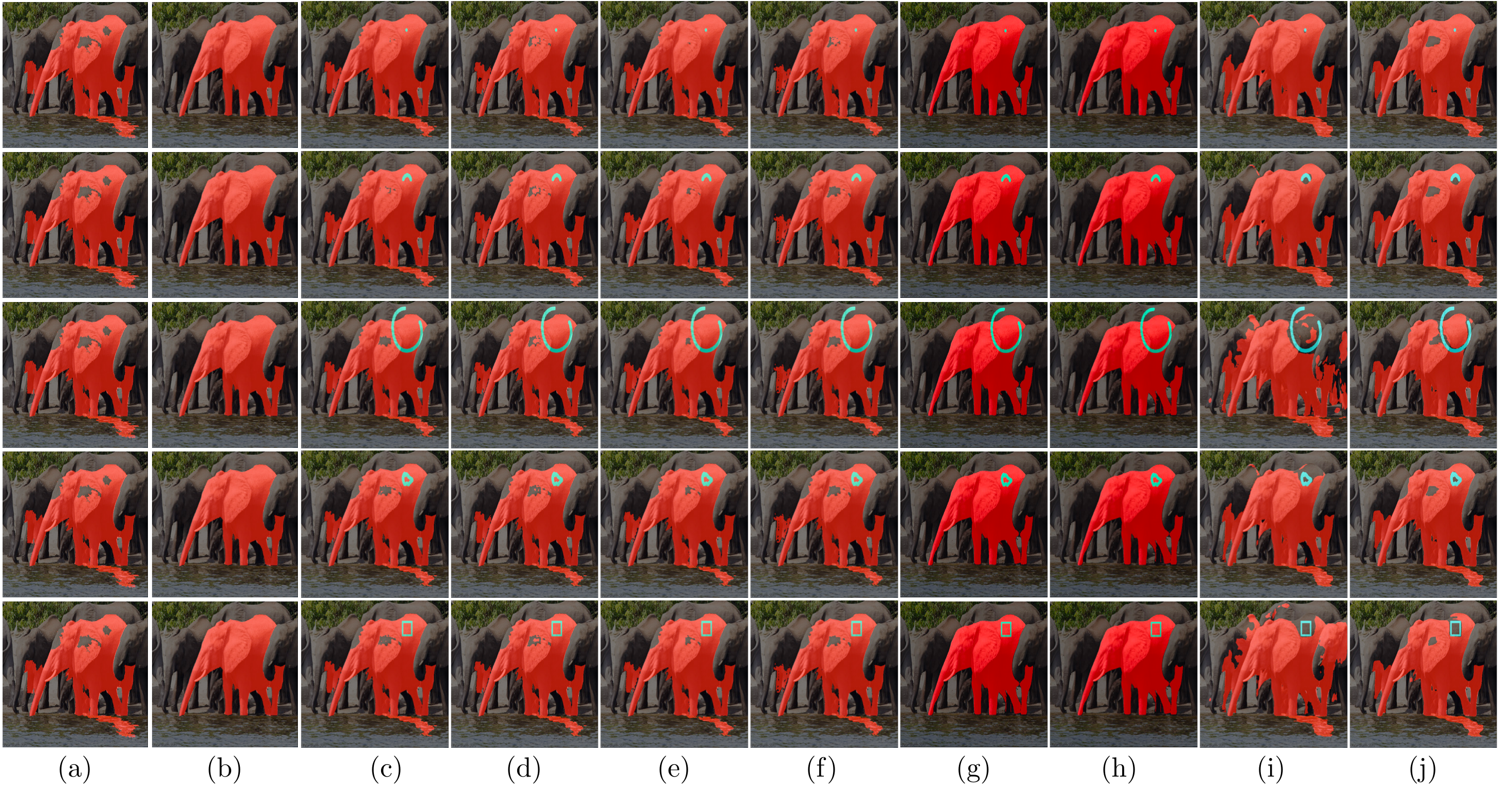}
    \caption{Qualitative results for each gesture type on DIG for segmentation refinement. (a) input image with previous segmentation overlayed, (b) input with with region ground truth overlayed (c) \emph{HRNet-multiHeadAug}, (d) \emph{HRNet-multiHead}, (e) \emph{HRNet-dataAug}, (f) \emph{HRNet-base}, (g) \emph{SAM~\cite{kirillov2023segment}-R - positive}, (h) \emph{SAM~\cite{kirillov2023segment}-C - positive},  (i) \emph{FocalClick~\cite{focalclick} - positive}, (j) \emph{RITM~\cite{reviving2021} - positive}.}
    \label{fig: refs}
\end{figure*}

\begin{figure*}[!bth]
    \centering
    \includegraphics[width=2\columnwidth]{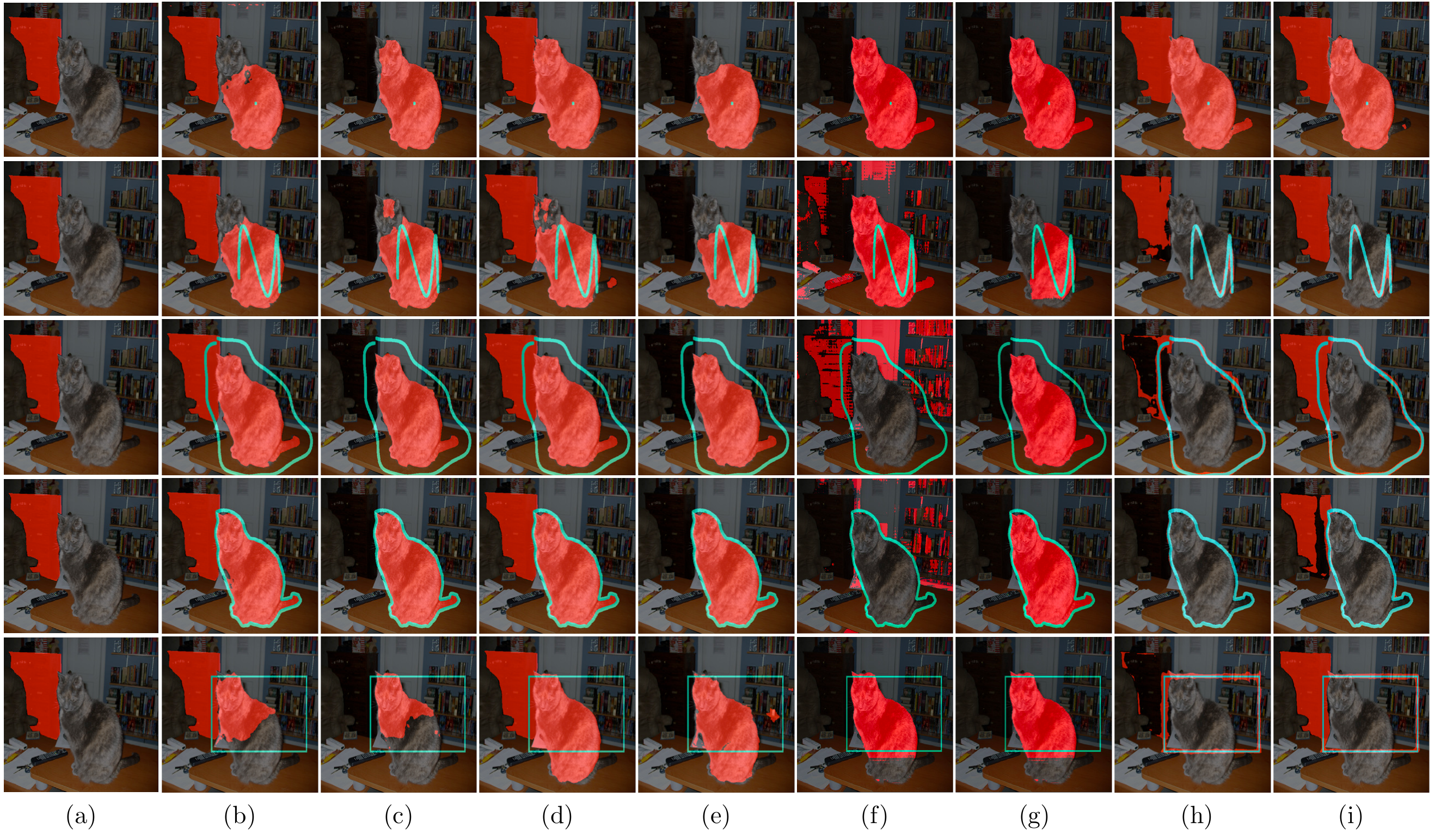}
    \caption{Qualitative results for each gesture type on DIG for multi-region segmentation. (a) input with with previous segmentation overlayed (b) \emph{HRNet-multiHeadAug}, (c) \emph{HRNet-multiHead}, (d) \emph{HRNet-dataAug}, (e) \emph{HRNet-base}, (f) \emph{SAM~\cite{kirillov2023segment}-R - positive}, (g) \emph{SAM~\cite{kirillov2023segment}-C - positive},  (h) \emph{FocalClick~\cite{focalclick} - positive}, (i) \emph{RITM~\cite{reviving2021} - positive}, (j) \emph{IOG~\cite{zhang2020interactive}}, (k) \emph{Deep GrabCut~\cite{xu2017deep}}.}
    \label{fig: muls}
\end{figure*}

\else
\begin{abstract}
Interactive segmentation entails a human marking an image to guide how a model either creates or edits a segmentation. Our work addresses limitations of existing methods: they either only support one gesture type for marking an image (e.g., either clicks or scribbles) or require knowledge of the gesture type being employed, and require specifying whether marked regions should be included versus excluded in the final segmentation.  We instead propose a simplified interactive segmentation task where a user only must mark an image, where the input can be of any gesture type without specifying the gesture type. We support this new task by introducing the first interactive segmentation dataset with multiple gesture types as well as a new evaluation metric capable of holistically evaluating interactive segmentation algorithms. We then analyze numerous interactive segmentation algorithms, including ones adapted for our novel task. While we observe promising performance overall, we also highlight areas for future improvement. To facilitate further extensions of this work, we publicly share our new dataset at \url{https://github.com/joshmyersdean/dig}. 

\end{abstract}
\section{Introduction}
A common goal is to locate regions, such as objects or object parts, in images.  We refer to this task as \emph{region segmentation}. A challenge is that fully-automated solutions often are error-prone while exclusive reliance on human annotations is costly and time-consuming. As a middle ground, \emph{interactive segmentation} methods empower humans to supply minimal input towards collecting \emph{consistently high-quality region segmentations}. Two popular interactive segmentation settings are to generate a segmentation (1) when the only input is a human marking on an image (i.e., \textbf{segmentation creation}) and (2) when the input is a human marking and a previous segmentation (i.e., \textbf{segmentation refinement}). 

\begin{figure}[!t]
    \centering
    \includegraphics[width=\columnwidth]{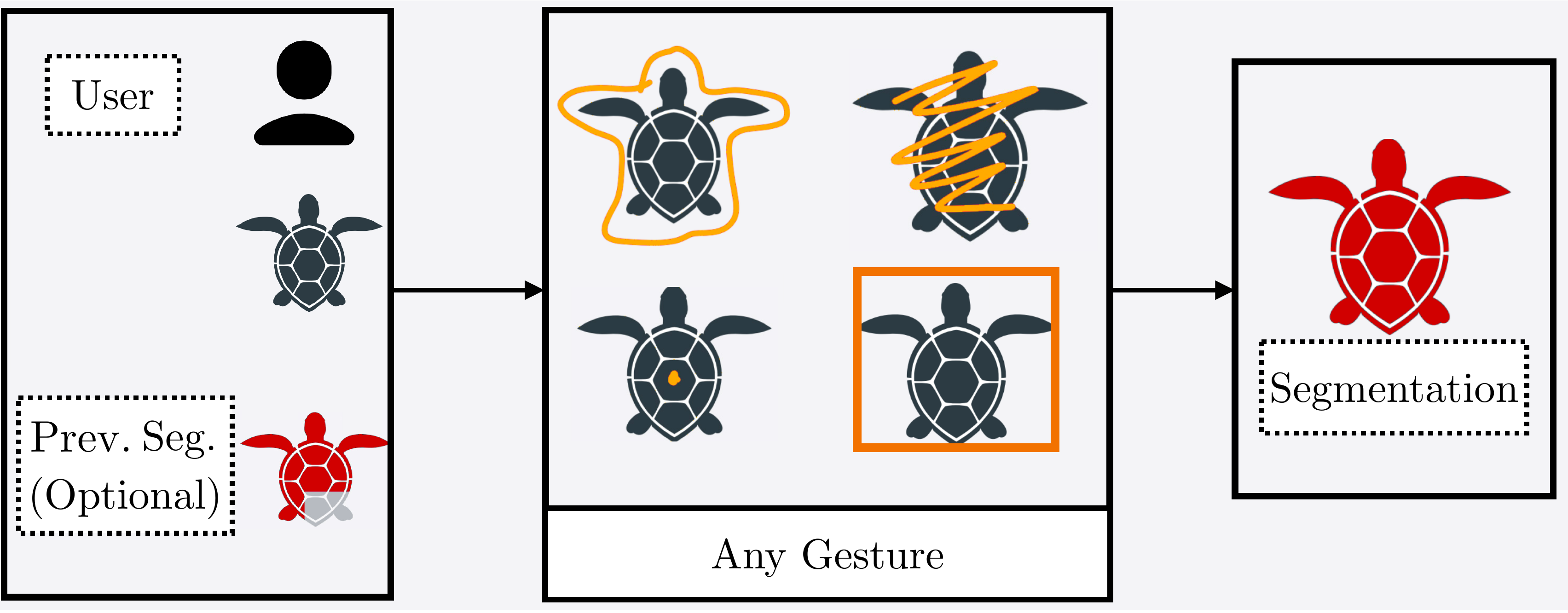}
    \vspace{-18pt}
    \caption{Overview of proposed interactive segmentation task. A user can mark an image with any gesture type which a method then uses to either create a segmentation from scratch or to refine a previous (imperfect) segmentation (if available).  This is done without any further guidance, including without specifying whether that marked region is content to include versus exclude.}
    \label{fig: pipe}
\end{figure}

While existing interactive region segmentation methods are widely-used and beneficial to society, they have at least one of the following two important limitations.  The first centers on how users interact with the methods. For most methods, a single \emph{gesture type} is supported as human input, such as either only clicks~\cite{focalclick, hao2021edgeflow, reviving2021, mahadevan2018iteratively, xu2016deep}, scribbles~\cite{bai2014error, grady2006random, li2004lazy, lin2016scribblesup, yang2010user}, lassos~\cite{xu2017deep, birkbeck2009interactive, shen2014interactive, peng2020deep, ling2019fast}, extreme points~\cite{maninis2018deep, roth2019weakly}, or rectangles~\cite{rother2004grabcut, xu2017deep, CHEN2018393, yu2017loosecut, zhang2020interactive}.  Yet, we will show findings from a user study in Section~\ref{sec: user_study} that users prefer different gestures in different scenarios.  The second limitation is that all methods, except a few that only can accept a single gesture type (e.g.,~\cite{xu2017deep}), require users to provide \emph{context} of whether the content they annotate should be included in or excluded from the final segmentation. This effort imposes an extra burden on humans that we hypothesize is unnecessary.

We propose a new interactive segmentation task where an algorithm \emph{only takes as input a gesture of any type}. An overview is shown in Figure~\ref{fig: pipe}.  This novel task relieves users from the burden of having to either learn how to use different tools for different gesture types or micro-manage how an algorithm interprets user input, such as signifying gesture type~\cite{kirillov2023segment, lin2022multi} or context of whether to include versus exclude the marked region~\cite{kirillov2023segment, lin2022multi, focalclick, reviving2021, xu2016deep, maninis2018deep}.  We call this novel task \emph{gesture-agnostic, context-free interactive segmentation}. 

We make several contributions in support of the novel task.  First, we present the first interactive segmentation dataset for explicitly training and evaluating models to support \emph{multiple gesture types}. We call it the \textbf{D}iverse \textbf{I}nteractive \textbf{G}esture (DIG) dataset and release it publicly to encourage community-wide progress. Second, we propose an evaluation metric to holistically evaluate algorithms for our new task. Finally, we benchmark the benefits of modern interactive segmentation algorithms for our proposed task, including after adapting their algorithmic frameworks for our task.  While we observe promising performance overall, we also highlight areas for future improvement. 

Success on this new task can yield numerous benefits to society. It would accelerate image editing, empowering users to simply use the gesture most natural to them when creating segmentations without becoming an expert on a vast array of `features' that support different gesture types and specifications for context. This could directly benefit lay users of image editing applications (e.g., Photoshop~\cite{adobephotoshop}) as well as specialized practitioners who depend on segmentations for downstream analysis (e.g., doctors performing diagnoses in the medical community~\cite{wang2018interactive, wang2018deepigeos, li2021wdtiseg}).  Such methods would also support more efficient curation of labeled training data to develop region segmentation models~\cite{acuna2018efficient, benenson2019large, castrejon2017annotating, kirillov2023segment}.  Finally, our work sets a precedent that could be generalized to segmentation tasks beyond region segmentation. For example, future studies could explore \emph{gesture-agnostic, context-free interactive segmentation} for tasks such as semantic segmentation~\cite{hu2019fully, wang2021pyramid, wang2021temporal} and panoptic segmentation~\cite{kirillov2019panoptic,liu2019end,mohan2021efficientps}. 

\vspace{-0.25em}\section{Related Work}
\paragraph{Interactions in Interactive Segmentation Methods.}
Interactive segmentation methods typically accept up to two types of information from users.  First, annotations of a region in an image are given to a model through gestures such as scribbles~\cite{bai2014error, grady2006random, li2004lazy, lin2016scribblesup, yang2010user, unal2022scribble}, lassos~\cite{xu2017deep, birkbeck2009interactive, shen2014interactive}, extreme points~\cite{maninis2018deep, roth2019weakly}, rectangles~\cite{rother2004grabcut, xu2017deep, CHEN2018393, yu2017loosecut, zhang2020interactive, kirillov2023segment}, and (most commonly) clicks~\cite{lin2020interactive, chen2021conditional, liew2017regional, koohbanani2020nuclick, majumder2020multi, kirillov2023segment}. For the subset of algorithms that can support multiple gesture types, they require additional input. For example, Segment Anything (SAM)~\cite{kirillov2023segment} requires contextual information during click interactions and specification of the gesture type to decide which prompt encoder (i.e., click, rectangle, or text) to utilize. Similarly, Multi-Mode Interactive Segmentation (MMIS)~\cite{lin2022multi} requires context as well as a bounding gesture (e.g., lasso, rectangle) for the initial segmentation before supporting non-bounding gestures (e.g., clicks). The second type of input common for interactive segmentation methods~\cite{castrejon2017annotating, mahadevan2018iteratively, focalclick, reviving2021, forte2020getting, kirillov2023segment} is context as to whether annotated content should be included or excluded in the final segmentation.  Only a few methods do not require this input.  Some assume marked content should always be included (e.g., Deep GrabCut~\cite{xu2017deep}). Alternatively, language can be used in place of gestures. While useful for many computer vision tasks~\cite{gurari_captions, liu2023llava} and potentially valuable in some interactive segmentation situations, relying only on language has important limitations. Not only do many vision-language models have poor multilingual support~\cite{luddecke2022image, carlsson2022cross, ding2020phraseclick, kazemzadeh2014referitgame}), but also often it can be difficult to articulate what one wishes to edit (e.g., a small correction to an existing segmentation). Extending prior work, we propose a simplified task of \emph{gesture-agnostic, context-free interactive segmentation}.  It is less cumbersome than the status quo as it reduces human effort to only a single input: marking an image with any gesture.

\vspace{-1em}
\paragraph{Interactive Segmentation Datasets.}
Most interactive segmentation methods are trained and/or evaluated on repurposed datasets originally designed for other tasks, such as semantic/instance segmentation~\cite{pascal, lin2014microsoft, gupta2019lvis, sbd, kuznetsova2020open, benenson2019large}, salient object segmentation~\cite{berk, rother2004grabcut}, entity segmentation~\cite{kirillov2023segment}, or video segmentation \cite{davis, focalclick}. For example, COCO+LVIS~\cite{reviving2021} combines two popular object segmentation datasets (i.e., COCO~\cite{lin2014microsoft} and LVIS~\cite{gupta2019lvis}) and is repurposed for interactive segmentation by removing semantic/instance information. DAVIS-585~\cite{focalclick} samples frames from the video segmentation dataset DAVIS~\cite{perazzi2016benchmark}, resulting in 300 images and 585 object segmentation annotations. The one exception is the SA-1B dataset~\cite{kirillov2023segment}, which was built using a large-scale human-machine collaboration to collect entity labels, resulting in 1.1 million images with 1 billion masks. Unlike existing datasets, we introduce the first dataset that comes with predefined gestures of different types and so enables developing one-size-fits-all algorithms to address the \emph{gesture-agnostic, context-free interactive segmentation} problem.

\vspace{-1em}\paragraph{Interactive Segmentation Evaluation Metrics.} 
The research community has employed a variety of evaluation metrics to assess the performance of interactive segmentation methods.  For example, some works rely on traditional measures for segmentation evaluation, including intersection-over-union (IoU)~\cite{xu2016deep, liew2021thin, maninis2018deep, lin2016scribblesup} and mean average precision (mAP)~\cite{xu2017deep}.  Alternatively, most click-based interactive segmentation methods~\cite{reviving2021, focalclick, lin2020interactive, lin2022focuscut, sofiiuk2020f} use the number of clicks (NoC), which captures how many clicks a model requires on average to achieve a target IoU.  A limitation of existing metrics is that, for methods that refine a previous segmentation, they fail to capture the extent to which resulting segmentations are better (or worse) than the previous segmentation.  We propose the first metric that can capture the relative change of a segmentation from a previous state, including an empty mask for segmentation creation.
\section{User Study on Gesture Type Preferences}
\label{sec: user_study}
We conducted a user study to establish what types of gestures people naturally gravitate towards when using interactive segmentation methods.  To our knowledge, no prior work has published such a study.

\vspace{-1em}\paragraph{Study design.} 
We designed 42 segmentation tasks reflecting a range of scenarios including segmenting salient objects, semantic categories, and multiple objects of different shapes (e.g., thin, occluded). Each segmentation task consisted of an image and short instruction at the top asking the participant to select something in the image. Participants were told, ``Use whatever gesture(s) that feel natural to you to make the selection asked: clicking, circling, drawing an outline, drawing a rectangle, drawing a polygon, scribbling, drawing a line, etc." To continue on to the next task, participants were instructed, ``After you are done making the gesture(s), you can move on to the next task. You will not receive any feedback from the app." Study participants were given the option to use either their fingers or a stylus when drawing a gesture. The study was conducted on an iPad with iPadOS 14 or above using an unbranded application in landscape mode with a scrollable and zoomable canvas.  

\vspace{-1em}\paragraph{Study participants.} 
We recruited participants from UserTesting\footnote{https://www.usertesting.com/} who identified as having either novice or intermediate photo editing skills.  In total, we had 43 participants.  Our participants reflected a range of demographics, including participants between the ages of 18 and 55 and a balanced gender ratio. Each participant was shown the 42 segmentation tasks in a randomized order. 

\begin{figure}[!t]
    \centering
    \includegraphics[width=\columnwidth]{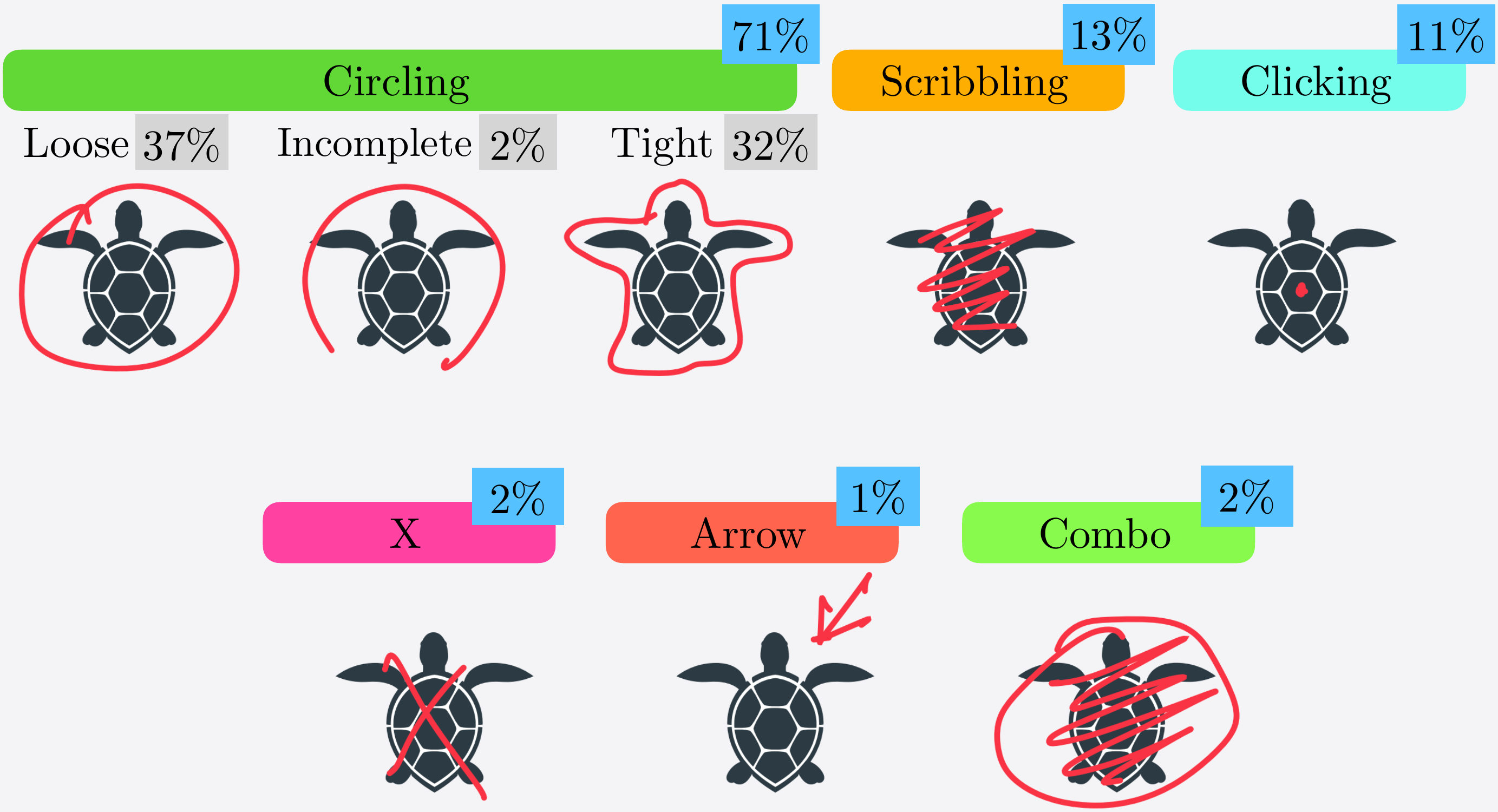}
    \vspace{-20pt}
    \caption{Breakdown of gesture frequency from our user study (N=1795) and examples of gestures. The observed diversity motivates designing algorithms to support multiple gesture types.}
    \label{fig: pie}
\end{figure}

\vspace{-1em}\paragraph{Results.} 
Overall, we collected 1795 gesture annotations (i.e., 42 segmentation tasks $\times$ 43 participants with some participants not completing all tasks). We tallied the type of gesture used based on the following gesture type categories: circling (i.e, lassos of varying granularity), scribbling, clicking, X's, arrows, and any combination of gestures. Gesture labels were assigned by visual inspection. Figure~\ref{fig: pie} shows the results of the user study.

We found that a diversity of gestures were used. Lassos were most common and varied between tight and loose. From further inspection, we found that different gestures were preferred in different contexts. For example, when selecting power lines in one image (i.e., a thin object based on visual inspection), participants used scribbles 54\% of the time and lassos 40\% of the time. In another task where participants were meant to select a chain link fence (i.e., another thin object), participants used scribbles 56\% of the time and lassos 26\% of the time. For another task, participants were instructed to select repeated sprinkles on a donut; participants used clicks 51\% of the time and lassos 28\%. In a task where participants needed to select a simple background (i.e., a wall with a woman in the foreground), participants used lassos 42\% of the time, clicking 26\% of the time, scribbles 21\% of the time, and X's or arrows 7\% of the time.

Overall, our findings underscore the value of one-size-fits-all interactive segmentation models that support a variety of gesture types to enable a more seamless and user-friendly experience for diverse users. Of note, great variation in gestures may also be observed on axes we did not test. For example, we only tested with an iPad. Gesture preference may vary if the user is using a mouse on a desktop, a trackpad on a laptop, or their finger on a small smartphone screen where their finger will occlude the target object. Variation may also exist across age groups or among people with motor or visual impairments~\cite{kane2011usable}.  This observation further motivates the need for our new approach to interactive segmentation, and we now discuss our work to establish this research direction.

\section{DIG Dataset}
We now present the \textbf{D}iverse \textbf{I}nteractive \textbf{G}esture (DIG) dataset, which supports two popular settings: generating a segmentation from only a marking on an image (i.e., \textbf{segmentation creation}) and generating a segmentation from an image, previous segmentation, and marking (i.e., \textbf{segmentation refinement}).  To our knowledge, DIG is the first interactive segmentation dataset with multiple gesture types. 

\subsection{Dataset Creation}
\label{subsec: creation}

\paragraph{Image Source.} 
We leverage 103,902 images from COCO+LVIS~\cite{reviving2021}, a popular source for interactive segmentation~\cite{reviving2021, focalclick, hao2021edgeflow}. Those images are advantageous because they each contain multiple objects, which in turn means that algorithms cannot simply learn to locate salient objects. An additional strength is that it includes a long tail of object types, a key motivation for creating LVIS~\cite{gupta2019lvis}.

\vspace{-1em}\paragraph{Ground Truth Region Segmentations.} 
We use the same source for the ground truth of region segmentations as our dataset source (i.e., COCO+LVIS~\cite{reviving2021}). Specifically, the ground truths are derived from instance-level segmentations. Consequently, we disregard semantic information since we are only concerned with locating a region within an image.  To generate ground truth for selecting \emph{part} of a region, we exploit that some regions in COCO+LVIS are broken into more than one region by occlusions. In such cases, we select one sub-region (i.e., connected component) as ground truth. 

\vspace{-1.1em}\paragraph{Dataset Filtering.} 
Following prior work on interactive segmentation datasets~\cite{focalclick}, we remove objects with an area of fewer than 300 pixels.  This acknowledges that such objects occupy a tiny portion of an image (e.g., at most 0.001\% of the area assuming a 512x512 resolution) and also that small areas are challenging for modern neural networks to identify due to information loss from operations such as downsampling and max pooling. Our final dataset consists of 103,902 images with 886,612 regions and 194,855 parts.

\vspace{-1.1em}\paragraph{Previous Segmentation Generation.} 
To enable fair, reproducible algorithm benchmarking for the interactive setting of refining a previous segmentation, we supply initial segmentations.  We employ the superpixel-based approach introduced in \cite{focalclick}.  For this interactive segmentation setting, we also consider segmenting \emph{multiple disconnected regions} (e.g., a banana and an orange sitting on a table).  We introduce this additional setting in order to facilitate teaching algorithms through training data to consider the relationship between the gesture and the underlying image content rather than only the gesture and the previous segmentation. Details can be found in the Supplemental Materials.

\vspace{-1.1em}\paragraph{Gesture Annotations.} 
We generate a static set of gesture annotations to enable consistent, fair algorithm comparison.\footnote{While one may consider generating gestures on-the-fly, this is not only inefficient but also impractical, as discussed in the Supplementary Materials.}  Motivated by our user study findings (Section~\ref{sec: user_study}), we focus on the following gestures most commonly observed in practice: lassos, scribbles, and clicks.  We also augment rectangles because they are a popular gesture in industry applications, such as the Rectangle Marquee tool in Photoshop~\cite{adobephotoshop}, and can lead to faster whole-object coverage than lassos~\cite{jain2013predicting}. We generate annotations corresponding to multiple gesture types for each region in our dataset, as exemplified in Figure~\ref{fig: scribbles}. A summary of how we construct each gesture is described below, and technical details are in the Supplementary Materials.

\begin{figure}[!t]
    \centering
    \includegraphics[width=\columnwidth]{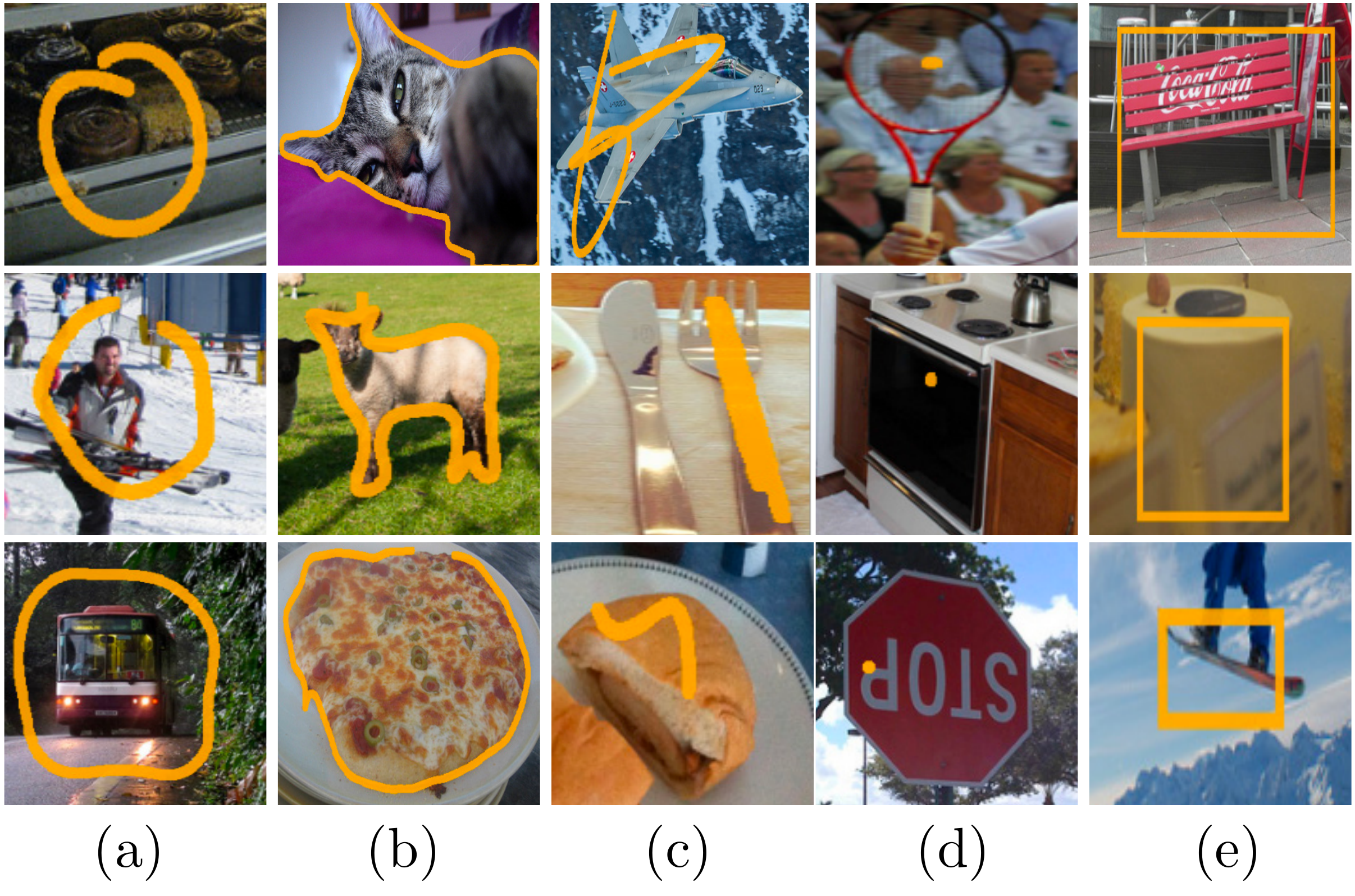}
    \vspace{-18pt}
    \caption{Examples of gesture types in DIG: (a) loose lassos, (b) tight lassos, (c) scribbles, (d) clicks, and (e) rectangles.  Images are cropped to the target region for visualization.}
    \label{fig: scribbles}
\end{figure}

For \textbf{lasso generation}, we create both loose and tight lasso gestures to accommodate variable user behaviors. For example, in some cases, a user may lack a steady hand, leading to imperfect or noisy lasso boundaries. We construct tight lassos by interpolating points sampled from a region boundary and ``jitter" these points to simulate user noise.

For \textbf{scribble generation}, we capture scribbles ranging from highly curved, squiggles to smooth curves. We randomly sample points from the target region or previous segmentation and then interpolate between them to create a B-spline curve followed by mechanisms to simplify curves and have scribbles pass outside the region's boundaries.

For \textbf{click generation}, we account for multiple click locations, spanning from each region's center to near its boundary.  Like prior work~\cite{xu2016deep}, we randomly select a foreground pixel (when creating a segmentation) or a pixel from a previous segmentation (when refining a segmentation).

\begin{table*}[!t]
\centering
    \resizebox{\textwidth}{!}{%
        \begin{tabular}{lcccccccc}
            \toprule
            \bf Dataset & \bf DIG & \bf DAVIS-585~\cite{focalclick} & \bf COCO+LVIS~\cite{reviving2021} & \bf GrabCut~\cite{rother2004grabcut, lempitsky2009image} & \bf Berkeley~\cite{berk} &  \bf PASCAL+SBD~\cite{pascal, sbd} & \bf OpenImages~\cite{benenson2019large, kuznetsova2020open} & \bf SA-1B~\cite{kirillov2023segment}\\ 
            \midrule
            \bf Gesture Types & L,C,S,R & - & - & R & - & - & - & -\\  
            \bf Prior Seg & $\checkmark$ & $\checkmark$ & \xmark & \xmark& \xmark& \xmark& \xmark & \xmark\\ 
            \bf \# Images & 104K & 300 & 104K & 50 & 300 & 11.5K & 1M & 11M\\
            \bf \# Regions & 1M & 585 & 1.5M & 50 & 300  & 31K & 2.8M & 1.1B\\
            \bf \# Samples & 13.5M & 585 & 1.5M & 50 & 300 & 31K  & 2.8M  & 1.1B\\ 
            \bottomrule
        \end{tabular}
    }
    \vspace{-8pt}
    \caption{Comparison of DIG to six interactive segmentation datasets. DIG is the only dataset to support multiple gesture types. We report the approximate number of samples rounded to the nearest order of magnitude. (L=Lasso, R=Rectangle, C=Click, S=Scribble)}
    \label{tab: compare}
\end{table*}

For \textbf{rectangle generation}, we use the approach presented by~\cite{xu2017deep}.  Given a tight bounding box, we modify the corners to perturb how closely the box encloses the region of interest. 

\subsection{Dataset Analysis}
We now characterize DIG and how it compares to seven existing interactive segmentation datasets: DAVIS-585~\cite{focalclick}, COCO+LVIS~\cite{reviving2021}, GrabCut~\cite{rother2004grabcut, lempitsky2009image}, Berkeley~\cite{berk}, PASCAL+SBD~\cite{pascal, sbd}, SA-1B~\cite{kirillov2023segment}, and OpenImages~\cite{benenson2019large, kuznetsova2020open}.  Those seven represent datasets that are long-standing interactive segmentation benchmarks for the research community (Berkeley, GrabCut), large-scale (COCO+LVIS, OpenImages, SA-1B), and popular for training interactive segmentation methods (COCO+LVIS, PASCAL+SBD, SA-1B).  For each dataset, we report the types of gestures included, whether previous segmentations for refinement are provided, number of images, number of unique regions, and total number of samples (i.e., number of unique gesture-region combinations).  Results are shown in Table~\ref{tab: compare}.

Our dataset is the only one providing pre-computed annotations of \emph{multiple gesture types}. The only other dataset supplying pre-computed interactive gestures is GrabCut~\cite{lempitsky2009image}, and only for `tight' rectangles (bounding boxes).  Consequently, our dataset is the first to support efficient training and evaluation of gesture-agnostic interactive segmentation algorithms by providing a standardized set of annotations that enables uniform comparison of algorithms.

Another distinction of our dataset is that it provides previous segmentations to support algorithm benchmarking for the segmentation refinement setting.  Only one other dataset supports this setting, DAVIS-585, but only with a testing split, thereby lacking splits for algorithm development. 

A further distinction is that DIG has at least four times as many samples as all datasets, except SA-1B~\cite{kirillov2023segment}. This arises primarily because our dataset includes multiple gesture types per each region in every image. We expect this to benefit deep learning methods (i.e., the de facto tool for interactive segmentation), which need large amounts of training data. While SA-1B contains more samples than DIG, it does not provide pre-computed previous segmentations or interaction annotations to enable consistent, and so fair, algorithm comparison.
\section{Evaluation Metric: RICE}
\label{sec: metric}
We now introduce a new evaluation metric for interactive segmentation to address that existing metrics (e.g., NoC, IoU) do not capture the amount that a method worsens/improves the results when refining an initial segmentation.  For example, they could not capture that a result \emph{worsens} when a segmentation has a lower IoU with the ground truth than a previous segmentation.  Similarly, existing metrics cannot distinguish a relatively smaller change from the previous segmentation compared to a large one such as when a model produces a prediction that has an IoU of 95\% when refining a previous segmentation with a high initial IoU (e.g., 90\%) versus a low initial IoU (e.g., 20\%).  We propose the first metric to holistically evaluate interactive segmentation algorithms regardless of gesture type and segmentation setting.

We call our metric the \textbf{R}elative \textbf{I}oU \textbf{C}orrective \textbf{E}valuation metric, or RICE. RICE takes into consideration how well a predicted segmentation improves/damages a previous segmentation with respect to a region's ground truth and simplifies to IoU when no previous segmentation is present.  Formally, we define RICE as:
\begin{align}
    \label{eqn: rice}
    RICE(\alpha, \beta) &= \begin{cases}
   \frac{\alpha - \beta}{1 - \beta}, & \text{if } \alpha \geq \beta\\
   \frac{\alpha}{\beta} -1,& \text{else}
   \end{cases},
\end{align}

\noindent
where $\alpha$ is the IoU between a region's ground truth and the output of an interactive segmentation model after thresholding. $\beta$ is the IoU between a previous segmentation and a region's ground truth.  Values can be positive or negative, where a negative value indicates there was a previous segmentation and, in the attempt to correct a mistake, the model prediction \emph{decreased} the overall IoU with the region ground truth rather than increasing it as desired.  Analysis highlighting RICE's derivation and intuition as well as its benefits over IoU are provided in the Supplementary Materials.   
\section{Algorithm Benchmarking}
\label{sec: methods}
We next benchmark modern interactive segmentation methods as is as well as adapted for our novel task. We perform all experiments on an NVIDIA A100 GPU. 

\vspace{-1em}\paragraph{Dataset Splits.} 
We leverage the splits used in the COCO+LVIS dataset~\cite{reviving2021} to divide DIG into training, validation, and test splits, since DIG is built upon that dataset.  We assign all training images from COCO+LVIS to our training dataset and then split the images in the validation set for COCO+LVIS into validation and testing splits using a random 70\%/30\% split. Our final dataset consists of a: \emph{training set} with 99,161 images, 857,669 regions, 186,740 region parts, and 12,839,936 samples; \emph{validation set} with 3,318 images, 32,938 regions, 5698 region parts, and 413,784 samples; and \emph{test set} with 1,423 images, 11,703 regions, 2417 region parts, and 246,080 samples. Of the 246,080 samples in the test set, 69,852 samples belong to the scenario of \emph{multi-region} segmentation, as described in Section~\ref{subsec: creation}.

\vspace{-1em}
\paragraph{Baseline Models.} 
Despite that algorithms do not exist to support our novel task, as all interactive segmentation methods require more input than our task permits, we still aim to gauge how well existing state-of-the-art interactive segmentation methods could work if modified for our task. We evaluate these top-performing algorithms that only support one gesture type since their code is publicly-available:

\begin{itemize}[itemsep=0.1em,leftmargin=*]

    \item \emph{Deep GrabCut~\cite{xu2017deep}:} top-performing model that takes as human input rectangles.  It supports segmentation creation.

    \item \emph{IOG:~\cite{zhang2020interactive}:} top-performing model that takes as human input a tight bounding box with a central click. It supports segmentation creation.\footnote{IOG~\cite{zhang2020interactive} refines its own predictions, not arbitrary previous ones, so it's unsuitable for comparison in our setting.} 

    \item \emph{RITM\cite{reviving2021}:} second best performing non-foundation model on four datasets for taking in human input click gestures. This method supports segmentation creation and refinement. We use the HRNet~\cite{hrnet}-18s variant. 

    \item \emph{FocalClick~\cite{focalclick}:} top performing non-foundation model on four datasets for taking as human input click gestures.  This method supports segmentation creation and refinement. We use the HRNet~\cite{hrnet}-18s variant.\footnote{We omit the refinement module as we observed worse performance with it; details are provided in the Supplementary Materials.} 
\end{itemize}
\noindent
To adapt these models for gesture-agnostic, \emph{context-free} interactive segmentation, as they do not explicitly support other gesture types, we convert all gesture types into the type supported by each model. For evaluation, since providing context to the models is incompatible with our proposed task, we instead assess each model using three approaches for supplying context:
\begin{itemize}[itemsep=0.05em,leftmargin=*]
\item  \emph{positive:} for each click, the method only receives content as `positive' (i.e., content should be included).
\item  \emph{negative:} for each click, the method only receives content as `negative' (i.e., content should not be included).
\item  \emph{random:} for each click, the method receives context decided by a draw from a discrete uniform distribution between positive and negative context.
\end{itemize}

\noindent
We also evaluate the state-of-the-art model that supports multiple gesture types, SAM~\cite{kirillov2023segment}.\footnote{MMIS~\cite{lin2022multi} is not evaluated since the code was not publicly available at the time of submitting this paper.}  Its official implementation\footnote{https://github.com/facebookresearch/segment-anything} supports clicks (context \emph{required}) and rectangles (context \emph{not required}). For other gesture types (i.e., scribbles and lassos), we either encode the points that compose the gesture as clicks (i.e., \emph{SAM-C}) or as rectangles (i.e., \emph{SAM-R}). We utilize the ViT-H~\cite{dosovitskiy2021an} variant of SAM and select the output mask (out of three) with the highest IoU with ground truth. 

\vspace{-1em}
\paragraph{Proposed Models.} 
We also introduce models that, by design, support multiple gesture types while only accepting as input gesture annotations.\footnote{For completeness, we also discuss two variants in the Supplementary Materials that highlight the performance of our adapted framework when permitting context. Overall, neither variant leads to a performance boost.} We adapt the HRNet~\cite{hrnet}-18s variant of FocalClick~\cite{focalclick} since it is both appropriate for use on a variety of devices~\cite{focalclick,reviving2021} because HRNet-18s is lightweight and also because it is the state-of-the-art for click-based segmentation outside of the parameter-heavy foundation models (e.g., SAM).  We introduce two variants, which we call \emph{HRNet-base}, \emph{HRNet-dataAug}.  Of note, these models will exemplify the advantage possible when training on our proposed DIG dataset.     

\emph{HRNet-base} is the original algorithm with two modifications. First, while FocalClick~\cite{focalclick} takes as input a concatenation of a channel for positive interactions (context), a channel for negative interactions (context), and a previous segmentation (that may be blank), we instead have it take as input a concatenation of the gesture, a Euclidean distance map, and a previous segmentation.  We exclude the positive and negative interactions because our problem does not permit the context of an interaction.  We introduce the Euclidean distance map because it has be shown to improve results when context is not present~\cite{xu2017deep}. 

\begin{table*}[!t]
  \begin{threeparttable}
    \resizebox{\textwidth}{!}{\begin{tabular}{ c l c c c c c c c c c c c c c c}
    \toprule
      & \multicolumn{1}{c}{}  & \multicolumn{2}{c}{\bf Average } &\multicolumn{2}{c}{\bf Click} & \multicolumn{2}{c}{\bf Scribble} & \multicolumn{2}{c}{\bf Loose Lasso} & \multicolumn{2}{c}{\bf Tight Lasso} & \multicolumn{2}{c}{\bf Rectangle} \\
     \cmidrule(lr){3-4}
     \cmidrule(lr){5-6}
     \cmidrule(lr){7-8}
     \cmidrule(lr){9-10}
     \cmidrule(lr){11-12}
     \cmidrule(lr){13-14}
         &\bf Method &  \bf RICE$_{\text{local}}$ & \bf RICE$_{\text{global}}$  & \bf RICE$_{\text{local}}$ & \bf RICE$_{\text{global}}$ & \bf RICE$_{\text{local}}$ & \bf RICE$_{\text{global}}$ & \bf RICE$_{\text{local}}$ & \bf RICE$_{\text{global}}$ & \bf RICE$_{\text{local}}$ & \bf RICE$_{\text{global}}$ & \bf RICE$_{\text{local}}$ & \bf RICE$_{\text{global}}$ & \\
        \midrule
        \multirow{14}{*}{\rotatebox[origin=c]{90}{\bf Creation}} 
        & \emph{RITM\cite{reviving2021} - positive} &29.83&28.90& 54.03&52.78&45.01&43.15&1.28&1.23&32.96&31.78&15.88&15.55 &\\ 
        & \emph{RITM\cite{reviving2021} - negative} & 10.68 & 10.45 & 0.00 & 0.00 &1.17 & 1.18 &11.23 & 11.23 &14.34 & 13.92 &26.64 & 25.94 &\\ 
        & \emph{RITM\cite{reviving2021} - random} & 20.33 & 19.74 & 27.15 & 26.47 &23.10 & 22.13 &6.32 & 6.30 &23.56 & 22.78 &21.49 & 21.00 &\\ \cmidrule(lr){2-16}
        & \emph{FocalClick~\cite{focalclick} - positive} &28.92&28.03& 54.95&53.47&44.78&43.12&1.29&1.26&29.49&28.44&14.11&13.86 &\\ 
        & \emph{FocalClick~\cite{focalclick} - negative} & 9.01 & 8.91 & 0.13 & 0.16 &0.70 & 0.74 &12.01 & 12.03 &9.92 & 9.87 &22.29 & 21.78 &\\ 
        & \emph{FocalClick~\cite{focalclick} - random} & 18.95 & 18.50 & 27.83 & 27.07 &22.86 & 22.02 &6.51 & 6.60 &19.38 & 19.07 &18.15 & 17.75 &\\ \cmidrule(lr){2-16}
        & \emph{SAM~\cite{kirillov2023segment}-R - positive} & \bf 67.25 & \bf 65.44 & \bf 77.94 & \bf 77.26 & 63.20 &  60.51 & 41.63 & 40.97 &74.25 & 71.83 & \bf 79.22 & \bf 76.63 &\\
        & \emph{SAM~\cite{kirillov2023segment}-R - negative} & 56.21 & 54.39 & 22.73 & 22.00 &63.20 & 60.51 &41.63 & 40.97 &74.25 & 71.83 &\bf 79.22 & \bf 76.63 &\\
        & \emph{SAM~\cite{kirillov2023segment}-R - random
        } & 61.75 & 59.93 & 50.42 & 49.70 &63.20 & 60.51 &41.63 & 40.97 &74.25 & 71.83 & \bf 79.22 & \bf 76.63 & \\\cmidrule(lr){2-16}
        & \emph{SAM~\cite{kirillov2023segment}-C - positive} & 55.54 & 54.39 &
    \bf 77.94 & \bf 77.26 &\bf 66.83 & \bf 64.85 &8.49 & 8.64 &45.20 & 44.55 & \bf 79.22 & \bf 76.63 &\\
        & \emph{SAM~\cite{kirillov2023segment}-C - negative} & 33.90 & 32.99 & 22.73 & 22.00 &20.78 & 20.13 &9.59 & 9.69 &37.19 & 36.52 & \bf 79.22 & \bf 76.63 &\\
        & \emph{SAM~\cite{kirillov2023segment}-C - random} & 44.67 & 43.63 & 50.03 & 49.36 &43.98 & 42.62 &9.05 & 9.18 &41.06 & 40.37 & \bf 79.22 & \bf 76.63 &\\\cmidrule(lr){2-16}
        & \emph{HRNet-base} &64.37&61.84& 54.85&52.02&59.55&56.99&67.24&65.66&80.88&77.25&59.33&57.29 &\\   
        & \emph{HRNet-dataAug} & 66.62&63.99&57.81&54.83& 61.25& 58.59& \bf 69.15&  \bf 67.51&  \bf 82.12& \bf 78.40& 62.79&60.65 &\\ \hdashline
        \multirow{14}{*}{\rotatebox[origin=c]{90}{\bf  Refinement}}
        & \emph{RITM\cite{reviving2021} - positive}& -18.59 & -11.13 & -1.49 & 10.15 &-3.74 & 5.84 &-43.84 & -40.61 &-24.50 & -17.89 &-19.36 & -13.13 &\\ 
        & \emph{RITM\cite{reviving2021} - negative}& -16.28 & -7.52 & -2.19 & 9.49 &-5.25 & 5.64 &-24.72 & -19.04 &-23.32 & -15.13 &-25.91 & -18.56 &\\ 
        & \emph{RITM\cite{reviving2021} - random}& -17.47 & -9.35 & -1.84 & 9.80 &-4.56 & 5.74 &-34.13 & -29.68 &-24.06 & -16.64 &-22.75 & -15.95 &\\ \cmidrule(lr){2-16}
        & \emph{FocalClick~\cite{focalclick} - positive} & -28.50 & -15.69 & -14.17 & 1.71 &-13.11 & 2.79 &-56.34 & -49.25 &-31.66 & -18.68 &-27.21 & -15.03 &\\ 
        & \emph{FocalClick~\cite{focalclick} - negative} & -24.56 & -11.68 & -7.28 & 11.85 &-14.62 & 1.16 &-36.89 & -29.41 &-30.23 & -18.70 &-33.79 & -23.29 &\\ 
        & \emph{FocalClick~\cite{focalclick} - random} & -26.55 & -13.76 & -10.70 & 6.78 &-13.89 & 1.83 &-46.80 & -39.60 &-30.83 & -18.58 &-30.51 & -19.23 &\\ \cmidrule(lr){2-16}
        & \emph{SAM~\cite{kirillov2023segment}-R - positive} & 20.95 & 21.31 & 42.40 & 44.22 &6.94 & 6.89 &25.79 & 25.92 &18.81 & 18.82 &10.81 & 10.71 &\\
        & \emph{SAM~\cite{kirillov2023segment}-R - negative} & -83.78 & -82.09 & -94.86 & -93.82 &-91.93 & -91.51 &-68.59 & -64.49 &-76.46 & -74.28 &-87.07 & -86.33 &\\
        & \emph{SAM~\cite{kirillov2023segment}-R - random
        } & 18.31 & 18.54 & 29.22 & 30.38 &6.94 & 6.89 &25.79 & 25.92 &18.81 & 18.82 &10.81 & 10.71 &\\ \cmidrule(lr){2-16}
        & \emph{SAM~\cite{kirillov2023segment}-C - positive}& -72.56 & -65.49 & -50.02 & -32.95 &-62.95 & -54.19 &-84.24 & -80.72 &-78.40 & -73.16 &-87.18 & -86.45 &\\
        & \emph{SAM~\cite{kirillov2023segment}-C - negative} & -90.42 & -89.20 & -94.86 & -93.82 &-92.42 & -90.86 &-88.40 & -87.11 &-89.21 & -87.78 &-87.18 & -86.45 &\\
        & \emph{SAM~\cite{kirillov2023segment}-C - random} & -81.51 & -77.35 & -72.49 & -63.34 &-77.85 & -72.69 &-86.19 & -83.74 &-83.84 & -80.51 &-87.18 & -86.45 &
\\ \cmidrule(lr){2-16}
        & \emph{HRNet-base} &36.94&44.84& 34.45&44.18&36.29&43.54&35.94&43.71&40.90&44.79&37.13&41.22 &\\   
        & \emph{HRNet-dataAug} &\bf 38.11&\bf 50.18& \bf 36.26& \bf 50.94&  \bf 37.91&  \bf 50.02& \bf 38.11&  \bf 51.44&  \bf 40.95&  \bf 51.06& \bf 37.33&\bf 48.22 &\\
        
    \bottomrule\\
    \end{tabular}}
    \end{threeparttable}
    \vspace{-15pt}
    \caption{Results on the test set of DIG. Above the dashed line represents segmentation creation and below represents segmentation refinement. SAM~\cite{kirillov2023segment} performs the best during segmentation creation while \emph{HRet-dataAug} the best during refinement.} \vspace{-1em}
    \label{table: results}
\end{table*}

\emph{HRNet-dataAug} is \emph{HRNet-base} with data augmentation to encourage algorithms to learn the relationship between a gesture and a region rather than a region alone.  To augment data, given an object with no previous segmentation, we set another image region as a previous segmentation and include it in the final ground truth with probability $p$, with $p=0.2$.  

\vspace{-1em}
\paragraph{Evaluation Metric.} 
We use our proposed RICE metric to evaluate both globally and locally.\footnote{To be backwards compatible in evaluation, we conduct an additional assessment that is discussed in the Supplementary Materials due to space constraints. We evaluate algorithms for three IoU thresholds on DAVIS585~\cite{focalclick} using the prior standard evaluation metric, number of clicks, adapted to number of gestures to accommodate multiple gesture types. Our findings show that context-augmented algorithms underperform and need more interactions compared to our proposed models.} The \emph{global} metric is defined to measure how well an algorithm's prediction matches the region's ground truth.  For our \emph{local} metric, we only consider how well the algorithm fixes a single connected mistake targeted by the gesture.\footnote{As outlined in the Supplementary Materials, our evaluation method is more realistic than previous approaches (e.g., ~\cite{lin2016scribblesup, xu2016deep, zhang2020interactive}). Unlike click-based methods that center clicks around the largest error, our interactions are varied in position, as shown in Figure ~\ref{fig: scribbles}.}  This is based in part on the observation that segmentation mistakes can occur on different parts of a region, causing multiple spatially disconnected groups of pixels that need to be corrected, and those disconnected mistakes will generally be corrected by users one at a time.  Similarly, for segmentation creation, the local metric establishes how well an algorithm selects a \emph{region part}, such as the head of a dog compared to it's whole body.

\vspace{-1em}
\paragraph{Experimental Design.} 
Given an interaction (i.e., click, scribble, lasso, rectangle), we examine the RICE score for that gesture in both segmentation settings (i.e., with a previous segmentation and without). Due to time and computation constraints, we do not consider combinations of multiple gestures for evaluation as there are ${5 \choose n}$ possible combinations for every region in the test set, where $n$ is the number of gestures chosen. An additional practical reason is that in an interactive session, the user will typically receive feedback from one marking before making the next marking. 

\vspace{-1em}
\paragraph{Overall Results.} 
Results are shown in Table~\ref{table: results}. We report a breakdown of these results, all results for poor performing models (i.e., Deep GrabCut~\cite{xu2017deep} and IOG~\cite{zhang2020interactive}), and multi-region segmentation results (which follow trends for our top-performing segmentation refinement method) in the Supplemental Materials.

As shown, models that support multiple gesture types outperform methods that support a single gesture type.  Within the top-performing models that support multiple gesture types simultaneously, we observe mixed outcomes.  For the segmentation creation setting, \emph{SAM-R - positive} (i.e., interactions mapped to rectangles with all context assumed to be positive) performs the best. However, this improvement comes at the cost that SAM-R requires the gesture type as input and is parameter-heavy.  In contrast, our simplified model with data augmentation (\emph{HRNet-dataAug}) achieves nearly comparable performance (i.e., 1.45 percentage points behind \emph{SAM-R - positive} with respect to RICE$_{\text{global}}$), while having 99.56\% fewer parameters and no extra input requirements.  For the segmentation refinement setting, \emph{HRNet-dataAug} performs the best overall. In contrast, SAM-C~\cite{kirillov2023segment} has the worst results, likely due to the problem that SAM discards the relevant mask when the first interaction has negative context. Overall, we contend that the advantages of the less cumbersome, more lightweight \emph{HRNet-dataAug} outweigh those of SAM based models.

For methods that only support clicks (i.e., RITM, FocalClick), not only do they require additional input information compared to our proposed models, but they also perform worse. For example, the most effective single-gesture approach (i.e., \emph{FocalClick - positive}) performs 2.86 percentage points worse than our baseline with respect to the RICE$_\text{local}$ score for clicks during segmentation creation. Similarly, it performs 1.36 percentage points worse with respect to RICE$_\text{global}$. We hypothesize this is partly because click-based methods are tailored for optimizing for the different metric of NoC for scenarios with multiple sequential interactions. 

Qualitative results are shown in Figure~\ref{fig: quals} for segmentation creation. Results for single-gesture methods (i.e., top two rows) reinforce the quantitative findings that they struggle to segment the region of interest. As exemplified, a plausible reason for their shortcomings is that gestures such as lassos and bounding boxes have the context outside or at the boundary of the region. While we observe a similar pattern for SAM~\cite{kirillov2023segment}-C, we find this issue is resolved when using instead rectangle encodings (i.e., SAM-R). This is likely because, unlike SAM-C, SAM-R does not rely on contextual information.  We also observe that in Figure~\ref{fig: quals}(b) that only the multi-gesture methods target the shirt of the baseball player rather than the entire player. We suspect this is due to the inclusion of annotations for region parts in the datasets used for training, including from our DIG dataset. A further encouraging outcome is that when the image marking targets the entirety of the desired region (i.e., using lassos), then the proposed appropriately segment the entire person. 

\begin{figure}[!t]
    \centering
    \includegraphics[width=.49\textwidth]{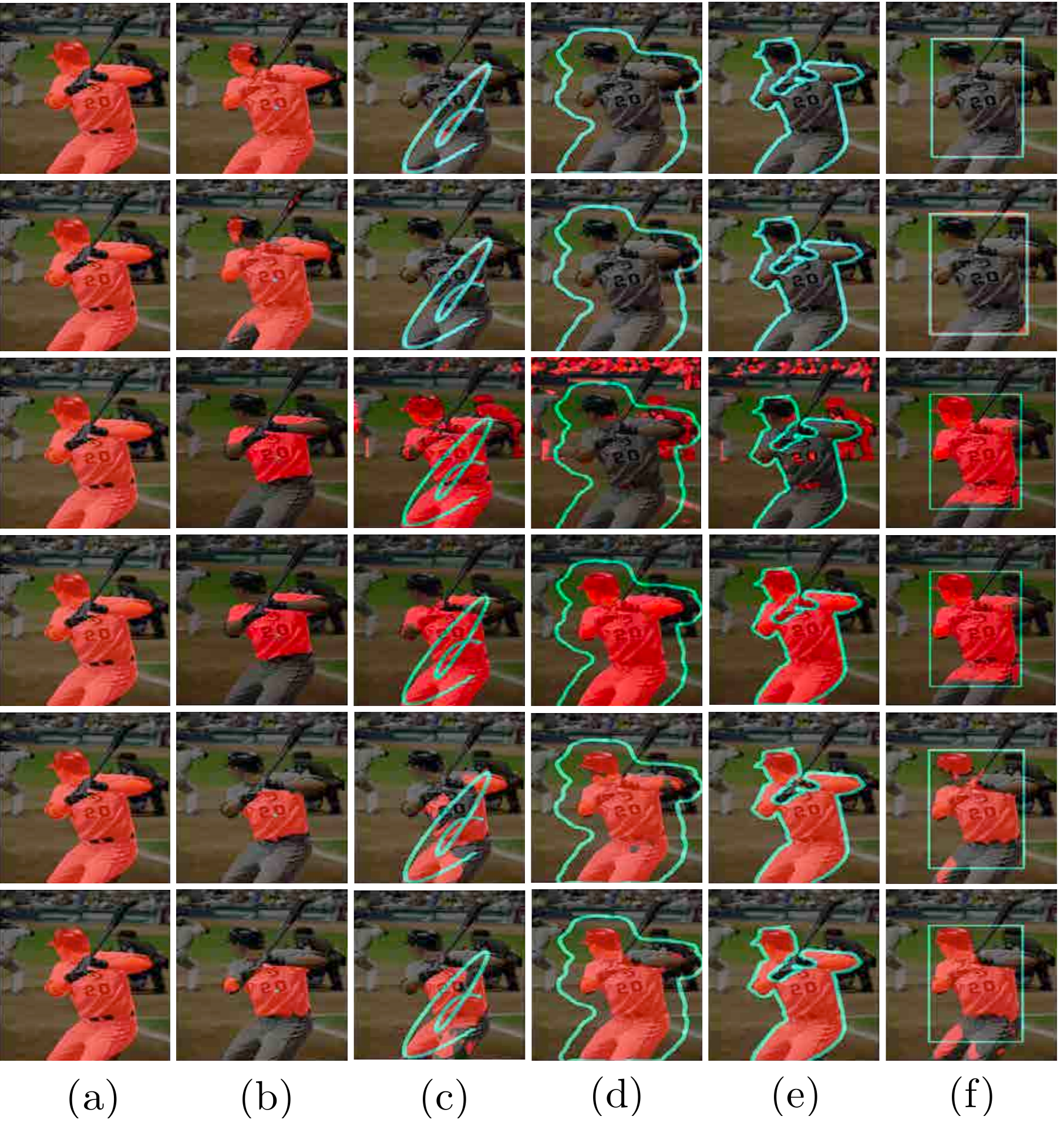}
    \caption {Results for each gesture type for segmentation creation. From the top down: RITM~\cite{reviving2021} - positive, FocalClick~\cite{focalclick} - positive, SAM~\cite{kirillov2023segment}-C - positive, SAM~\cite{kirillov2023segment}-R - positive, HRNet-base, HRNet-dataAug. (a) is the input image with ground truth overlaid, (b)-(f) show the results of each method with the gesture used.}
    \label{fig: quals}
\end{figure}

\vspace{-1em}\paragraph{Analysis With Respect to Gesture Type.}
There is a disparity in the performance of different gesture types across methods. Among \emph{single-gesture methods}, clicks yield the most favorable results, while loose lassos display the least effectiveness. This observation is likely due to the training approach, where the interactions in single-gesture methods are typically centered around the region of interest. In contrast, loose lassos typically fall outside the region of interest, although they may intersect with it due to the boundary sampling and interpolation methods discussed in Section~\ref{subsec: creation}. For \emph{multi-gesture} methods such as SAM~\cite{kirillov2023segment}, we find that clicks and rectangles yield similar results during segmentation creation with positive context, likely due to the fact that these are the gestures that are natively supported by SAM, requiring no separate encoding.

In contrast to methods that take in context, our HRNet variants show better performance with tight lassos across all evaluation metrics, while clicks tend to yield poorer results during segmentation creation. Intuitively, a tight lasso surrounding the region's boundary provides the most guidance on what to select for interactive segmentation methods while clicks perform the least. Scribbles and rectangles provide similar performance as they may both only envelope \emph{part} of a region of interest. However, we observe that SAM obtains top performance for segmentation creation when using rectangles. This can likely be attributed to rectangles being the supported gesture type that provides the most guidance for SAM during its large-scale training.

We also observe that the disparity in performance between different gesture types is smaller for segmentation refinement (Table~\ref{table: results}). For instance, when using \emph{HRNet-dataAug}, the gap between the RICE$_\text{local}$ score achieved by clicks and tight lassos reduces from 24.31 percentage points for segmentation creation to 4.96 percentage points during refinement. One plausible explanation is that the refined segments are typically smaller in size than the entire regions, thereby allowing algorithms to respond more uniformly among gesture types. However, as the spatial size of available corrections diminishes, the utilization of clicks becomes increasingly advantageous. In the NoG setting, described in the Supplemental Materials, we observe that clicks outperform other methods in terms of minimizing failures in reaching a specified IoU. This may be attributed to corrections becoming thinner as they become smaller. Consequently, the effectiveness of boundary level guidance, especially when applied with a fixed thickness (e.g., a radius of 5 in our annotations), may be diminished. In contrast, clicks may be more desirable due to their ability to cover a smaller spatial extent. An additional explanation is that the observed performance advantage of clicks during refinement could be influenced by the implicit bias from training with superpixel previous segmentations in DIG. Future research could explore more efficient methods for generating on-the-fly interactions to alleviate potential biases.

\vspace{-1em}\paragraph{Analysis with Respect to Segmentation Refinement.}
Our results on segmentation refinement reveal that single-gesture methods have limited ability to improve upon previous segmentations, while multi-gesture methods display comparatively better but still suboptimal performance. We observe that context-based methods struggle to enhance previous segmentations when using a single interaction, as evidenced by our proposed RICE metric. One possible explanation is that the widely used metric of IoU may not be a reliable indicator of how well a method has improved a prior segmentation, if at all. For example, when analyzing the RICE$_\text{global}$ score for the RITM method, we observe a relatively low score of 6.11, despite achieving a mean Intersection over Union (mIoU) of 83.74 for the same setting\footnote{Due to space constraints, we report IoU for each method in the Supplementary Materials.}. Moreover, methods that rely on interaction history, such as SAM perform poorly when correcting pre-computed segmentations due to the requirement for knowledge of previous interactions. Under the NoG setting, we find this issue remedied by leveraging subsequent interactions, but find these algorithms struggle when context is not available. Furthermore, SAM-R suffers a disadvantage when refining a previous segmentation as it expects rectangles to \emph{add} content to a segmentation, rather than remove it.

\section{Conclusion}
Our proposed \emph{gesture-agnostic, context-free} interactive segmentation task suports a less cumbersome, more flexible interaction from  users. By only accepting user markings on images, it eliminates common additional input requirements, such as the context of an interaction or type of gesture. 

\fi

\vspace{0.5em}\noindent\textbf{Acknowledgments.} JMD is supported by a NSF GRFP fellowship under Grant No. 1917573 and completed the majority of this work during an internship with Adobe Research. We gratefully thank the participants of our user study, members of the Media Intelligence Lab at Adobe Research for early feedback, and Eric Slyman for helpful discussions about constructing previous segmentations. We also thank the anonymous reviewers for their feedback.

\ifsupp
\clearpage
\fi
{\small
\bibliographystyle{ieee_fullname}
\bibliography{egbib}
}

\end{document}